\documentclass[10pt,journal]{IEEEtran}
\usepackage{graphicx}
\usepackage{amsmath,amsfonts}
\usepackage[T1]{fontenc}
\usepackage{times}
\usepackage{algorithmic}
\usepackage{algorithm}
\usepackage{array}
\usepackage[caption=false,font=normalsize,labelfont=sf,textfont=sf]{subfig}
\usepackage{textcomp}
\usepackage{stfloats}
\usepackage{url}
\usepackage{verbatim}
\usepackage{cite}
\usepackage{multirow}
\usepackage{subfloat}
\usepackage{tabularray}
\UseTblrLibrary{booktabs}
\usepackage[switch]{lineno}
\usepackage[svgnames, x11names]{xcolor}
\usepackage{hyperref}
\hypersetup{
    colorlinks=true,
    urlcolor=RoyalBlue,
    citecolor=RoyalBlue,
    linkcolor=RoyalBlue,
}

\begin{document}

\title{Leveraging Summary Guidance on Medical Report Summarization}
\author{
    Yunqi Zhu, Xuebing Yang, Yuanyuan Wu, and Wensheng Zhang
    \thanks{
        Corresponding authors: Xuebing Yang; Wensheng Zhang.
    }
    \thanks{
        Yunqi Zhu and Wensheng Zhang are with the School of Information and Communication Engineering, Hainan University, Haikou, 570228, China, and also with the Institute of Automation, Chinese Academy of Sciences, Beijing 100190, China (email: zhuyunqi96@163.com; zhangwenshengia@hotmail.com).
    }
    \thanks{
        Xuebing Yang is with the Institute of Automation, Chinese Academy of Sciences, Beijing 100190, China (e-mail: yangxuebing2013@ia.ac.cn).
    }
    \thanks{
        Yuanyuan Wu is with the School of Information and Communication Engineering, Hainan University, Haikou, 570228, China (e-mail: wyuanyuan82@163.com).
    }
    \thanks{
        The code is available at: \url{https://github.com/zhuyunqi96/medreportsum}
    }
}

\markboth{Preprint}{}


\maketitle
\begin{abstract}
This study presents three deidentified large medical text datasets, 
named DISCHARGE, ECHO and RADIOLOGY, 
which contain 50K, 16K and 378K pairs of report and summary that are derived from MIMIC-III, respectively.
We implement convincing baselines of automated abstractive summarization on the proposed datasets with pre-trained encoder-decoder language models, including BERT2BERT, T5-large and BART.
Further, based on the BART model, we leverage the sampled summaries from the train set as prior knowledge guidance, 
for encoding additional contextual representations of the guidance with the encoder and enhancing the decoding representations in the decoder.
The experimental results confirm the improvement of ROUGE scores and BERTScore made by the proposed method, outperforming the larger model T5-large.

\end{abstract}


\section{Introduction}

\IEEEPARstart{A}{utomated}
medical report summarization is a critical generation task for efficiently producing concise summaries from lengthy documents such as medical reports\cite{standford-sum, multi-doc-sum, re3writer-model} and articles\cite{aceso-art-sum, sumpubmed, pretrainbiomed-art-sum}.
In recent years, there has been a growing interest in the research of free-form natural language generation in the medical domain, 
including medical text generation and medical image captioning\cite{med-image-cap-baseline, med-image-cap-KER, med-image-cap-topics, med-image-cap-ppked, med-image-cap-priorguided, med-image-cap-vlpretrain}.
Derived from MIMIC-III (a large public database with 50K deidentified hospitalization records), Shing et al.\cite{multi-doc-sum} proposed a dataset with 6K pairs of admission note and discharge summary, the collected report-summary pairs were split into seven subsets (Chief Complaint, Family History, Social History, Medications on Admission, Past Medical History, History of Present Illness and Brief Hospital Course) for independent summarization.
Furthermore, extracted from MIMIC-III, Liu et al.\cite{re3writer-model} proposed a dataset ``PI'' (Patient Instruction), containing 35K pairs of hospitalization record and discharge instruction.

To enhance the feasibility and accessibility of medical report summarization, 
this study collected three datasets from MIMIC-III, 
named DISCHARGE, ECHO and RADIOLOGY, 
containing 50K, 16K and 378K pairs of medical report-summary respectively: 
a) DISCHARGE contains holistic hospitalization records of the patient, 
with corresponding free-form discharge diagnosis summary; 
b) ECHO includes detailed echocardiography of the patient, 
with corresponding abstractive impression as the summary; 
c) RADIOLOGY includes elaborate radiography findings as the source document and free-form impression as the summary.

This study focuses on abstractive summarization on the medical report. 
Abstractive summarization is usualy considered as a sequence-to-sequence (Seq2Seq) natural language generation problem in machine learning that requires that the model automatically generate a shorter version of a longer text while preserving the overall meanings.
We fine-tune the proposed datasets with pre-trained Seq2Seq language models BERT2BERT, T5-large and BART, 
These models use an encoder-decoder framework, 
in which the encoder projects the source document to generate high-dimensional hidden representations, 
which are then used by the decoder to output the summary.
Further, for each of the proposed medical report datasets, 
we statistically and intuitively show that the hidden topical semantics representations of the summaries in the train set could share strong characteristics, making it difficult to cluster the summaries into multiple groups.
Next, based on the pre-trained model BART, we leverage the summaries in the train set, 
concretely sample a summary as prior knowledge guidance, 
for encoding additional semantic representations of the guidance with the encoder and enhancing the decoding representations in the decoder.
Finally, the experimental results show that the proposed method decently improves the summary quality and outperforms larger language model T5-large in ROUGE-1, ROUGE-L and BERTScore.

The main contributions of this study can be summarized as follows: 
\begin{itemize}
    \item We collect three medical report summarization datasets DISCHARGE, ECHO and RADIOLOGY from MIMIC-III,
            containing 50K, 16K and 378K pairs of deidentified hospitalization report-summary respectively.
    \item Strong baselines of abstractive summarization with pre-trained language models BERT2BERT, T5-large and BART are given.
    \item We leverage the sampled summary from the train set of the dataset as the summary guidance for improving the representation learning in the encoder-decoder framework.
            Evaluating the baseline models and the proposed method with ROUGE, BERTScore, SummaC and QuestEval, experimental results support the improvement of the proposed method.
\end{itemize}

\section{Background and Related Work}

\subsection{Abstractive Summarization}
Automated text summarization is a natural language generation task that condenses a long document into a succinct one while preserving the most significant information\cite{ai-review-sum-survey, aaai-sum-survey, eswa-sum-survey}.
Typically, there are two types of text summarization: abstractive and extractive.
Abstractive summarization is an important task that generates free-form summaries by compressing the contexts and rewriting the phrases as human would do,
the task is often phrased as a Seq2Seq language modeling problem in machine learning.
The purpose of extractive summarization is to highlight a limited number of significant sentences and then relocate the selected sentences as the summary,
hence the extractive summarization could be considered as a sentence-level multi-label classification problem in machine learning.

Recent abstractive summarization researches pay great attention to Seq2Seq language models, 
pre-training a large transformer-based\cite{transformer-model} model with masked language modeling\cite{bert-model, t5-model, unilm-model, bart-model} 
and then fine-tuning the model to downstream tasks have successfully improved the performance of text generation problems 
such as abstractive summarization\cite{bertabsext, pegasus-model, prophetnet-model}, machine translation and question answering.
State-of-the-art abstractive summarization systems tried bringing factual knowledges\cite{post-edit-correct-fact, fes-model, factpegasus-model} and extractive summary representations\cite{gsum-model, salience-ext-alloc-model} into the abstractive summarization system.
GSUM\cite{gsum-model} is an encoder-decoder framework that can utilize another summarization method (e.g. a fine-tuned summarization model\cite{bertabsext, bart-model}) to generate a candidate summary as a summary guidance.
Apart from the original encoded sequence embeddings from the source document, the framework gets the sequence embeddings of the guidance through the shared parameters encoder, and then do additional cross-attention with the decoding sequence in every decoder layer of the decoder.
Moreover, the evaluation of machine-generated summary mainly focuses on conventional overlapping degree of n-grams (i.e., ROUGE \cite{rouge-metric}), 
language model based semantic embeddings (e.g. BERTScore\cite{bertscore-metric} and BARTScore\cite{bartscore-metric}),
and factual consistency between the source and the summary (e.g., SummaC\cite{summac-metric}, QuestEval\cite{questeval-metric} and FactCC\cite{factcc-metric}).

\subsection{Medical report summarization}
Moen et al.\cite{clinical-sum-sys} proposed an extractive electronic health record summarization system with semantic-based models,
and Lee\cite{ehr-sum-sys} used recurrent neural network for abstractive summarization on electronic health record.
Further, Zhang et al.\cite{standford-sum} collected radiology summarization dataset with 87K reports from Standford Hospital, though the dataset is not publicly available.
Then, Zhang et al.\cite{standford-sum-opt-fact} improved the above dataset with fact-guided reinforcement learning strategy.
Next, mined from MIMIC-III, Shing et al.\cite{multi-doc-sum} implemented abstractive summarization and named entity recognition over 6K pairs of admission report and discharge summary, the report-summary pairs were further split into seven subsets (Chief Complaint, Family History, Social History, Medications on Admission, Past Medical History, History of Present Illness and Brief Hospital Course) for independent summarization,
Further, Liu et al.\cite{re3writer-model} proposed dataset ``PI'' and model Re$^3$Writer, 
PI (Patient Instruction) is a summarization dataset collected from MIMIC-III,
the dataset has 35K pairs of medical report and discharge instruction.
While Re$^3$Writer extends the transformer-based encoder-decoder model with additional top-K related instructions retrieval module and medical knowledge encoding module. 
The instructions retrieval module requires the storage of input embeddings of the train set as well as the computation of cosine similarity between the encoded embeddings of the input medical report and each of those in the training corpus, 
which can bring about large latency.

\section{Methodology}

\subsection{Datasets Collection}

We collected three medical report summarization datasets from MIMIC-III (Medical Information Mart for Intensive Care) \cite{mimic-iii, mimic-iii-paper}.
MIMIC-III is a publicly available single-center large medical database with about 50K deidentified hospitalization records admitted to critical care units between 2001 and 2012,
the data involves sections of medication, treatment, diagnosis, discharge report etc.
We mine and clean the free-form notes with basic string pattern matching over the section titles,
Note that the collected datasets have only one modality (i.e., text),
and both the source document and the summary are free-form notes. 
In this paper, we refer the datasets as DISCHARGE, ECHO and RADIOLOGY:
\begin{itemize}
    \item[1)] DISCHARGE contains a total of 50,258 discharge report-summary pairs;
            The report can include the free-form notes of admission, history of medical treatment, nursing observation, pharmacy etc.,
            while the final diagnostic summary will outline critical diagnoses of the patient during the hospitalization.
    \item[2)] ECHO has a total of 16,245 echocardiography report-summary pairs;
            The report will outline findings on the patient, 
            and the summary will highlight a few critical diagnostic results.
    \item[3)] RADIOLOGY contains a total of 378,745 radiography report-summary pairs;
            The physician will discuss detailed findings on the report and then summarize a few significant impressions of the inspection result.
\end{itemize}

Statistics of the datasets are shown in Table~\ref{tab:stats_dataset}, 
the dataset are randomly split into 80\%/10\%/10\%.
The average number of words and sentences are counted with the NLTK\footnote{\url{https://pypi.org/project/nltk}} package.
Although the level of novel bi-gram in the summary of DISCHARGE is slightly lower than the other two datasets, 
DISCHARGE has a higher level of text compression ratio between the source and summary than the others.
Further, ECHO and RADIOLOGY have relatively shorter source length, 
they may have less difficulty in compressing the overall contextual representations than DISCHARGE.
Note that the dataset ``PI'' (Patient Instruction), proposed by Liu et al.\cite{re3writer-model}, has 35K pairs of hospitalization record and discharge instruction that was also extracted from MIMIC-III.
Our DISCHARGE has similar input with PI, but the target output of DISCHARGE is the discharge summary that involves highly abstractive diagnoses.

\begin{table}[H]
\centering
\renewcommand\arraystretch{1.1}
\scalebox{0.83}{
    \begin{tblr}
    {   colspec=ccccccc,
        cell{1}{1}={r=2}{},
        cell{1}{2}={r=2}{},
        cell{1}{3,5}={c=2}{},
    }
    \hline
    Dataset &
    Train/Eval/Test &
    Avg. Source & &
    Avg. Summary & &
    $\%$ novel 
    \\
    \cmidrule[lr]{3-4} \cmidrule[lr]{5-6}
      &  & Words & Sents & Words & Sents & bi-gram
    \\ 
    \hline
    DISCHARGE & 40K/5K/5K &
    2162.20 & 100.10 & 28.84 & 2.21 & 60.52
    \\
    ECHO & 13K/1.6K/1.6K &
    \, 315.30 & \, 30.87 & 49.99 & 4.16 & 70.07
    \\
    RADIOLOGY & 300K/37K/37K &
    \, 168.49 & \, 11.18 & 46.09 & 2.94 & 72.76
    \\ \hline \\
    \end{tblr}
}
\caption{Statistics of the datasets.
$\%$ novel bi-gram indicates the average ratio of bi-grams that the reference summary contains,
while the corresponding source document does not.
}
\label{tab:stats_dataset}
\end{table}

\subsection{Summary Guidance on fine-tuning}

\begin{figure}[t]
\centering
\subfloat[\small{K $=$ 2}]{
\includegraphics[width=0.225\linewidth]{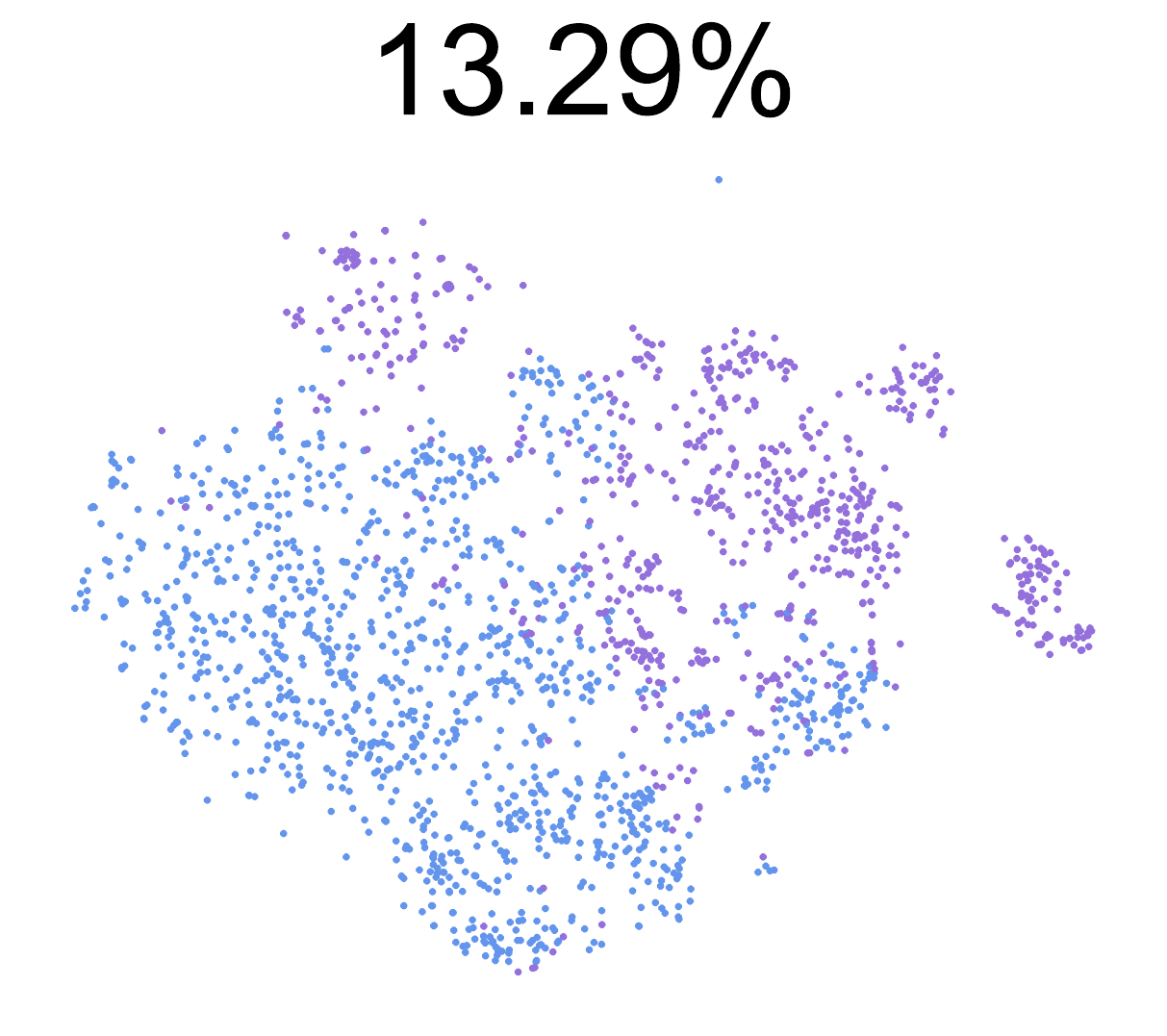}
}
\subfloat[\small{K $=$ 4}]{
\includegraphics[width=0.225\linewidth]{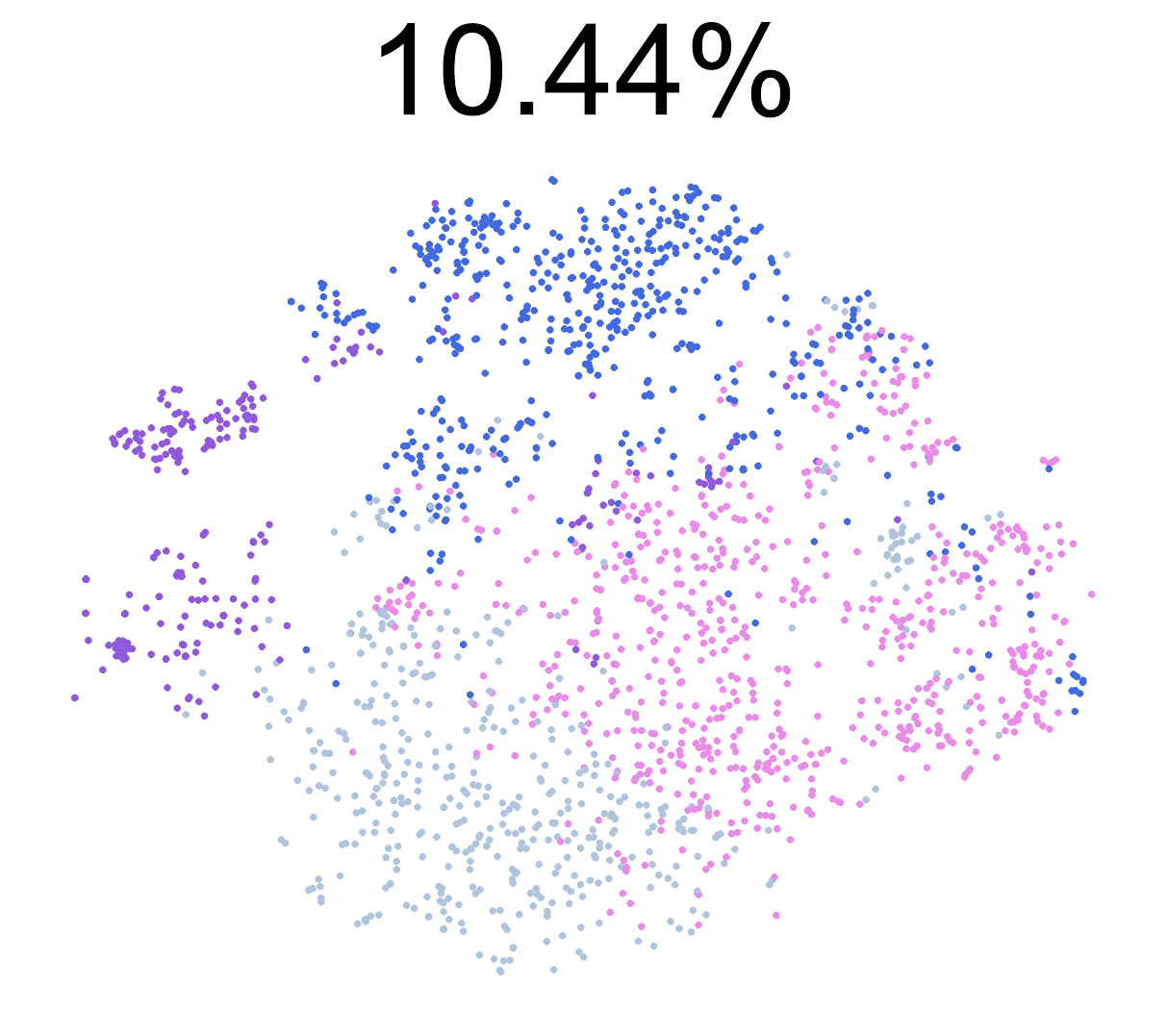}
}
\subfloat[\small{K $=$ 8}]{
\includegraphics[width=0.225\linewidth]{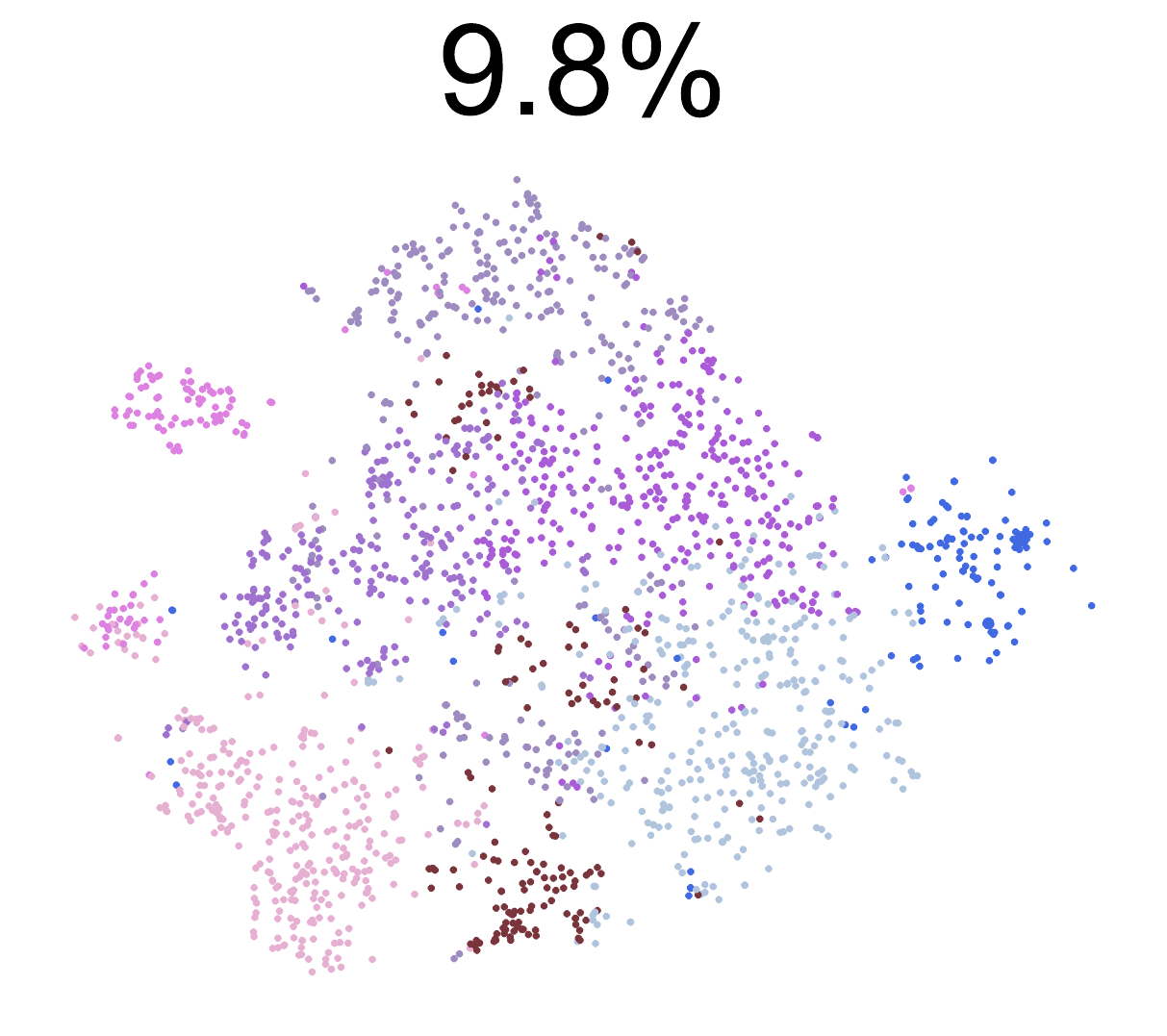}
}
\subfloat[\small{K $=$ 16}]{
\includegraphics[width=0.225\linewidth]{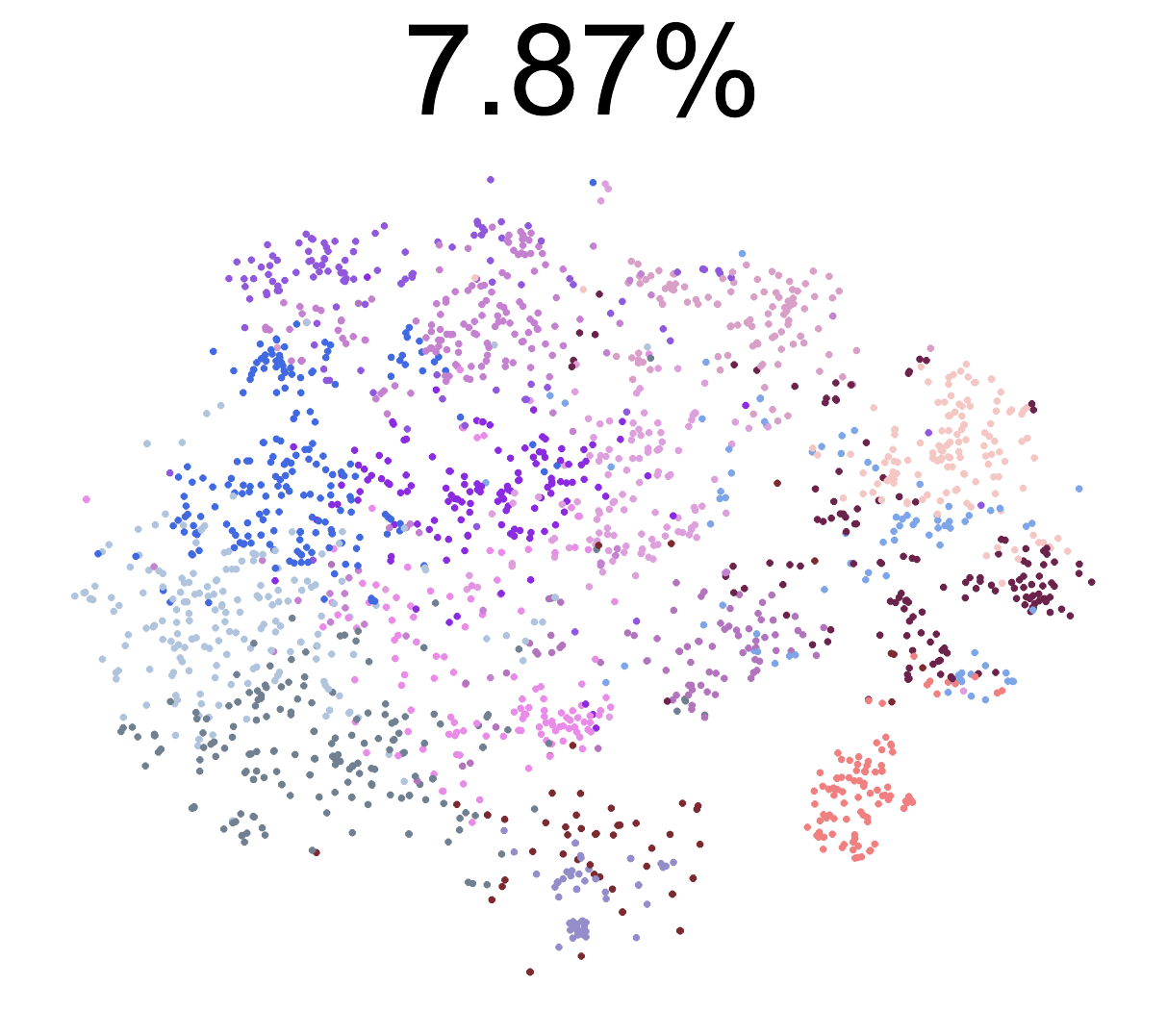}
}

\subfloat[\small{K $=$ 2}]{
\includegraphics[width=0.225\linewidth]{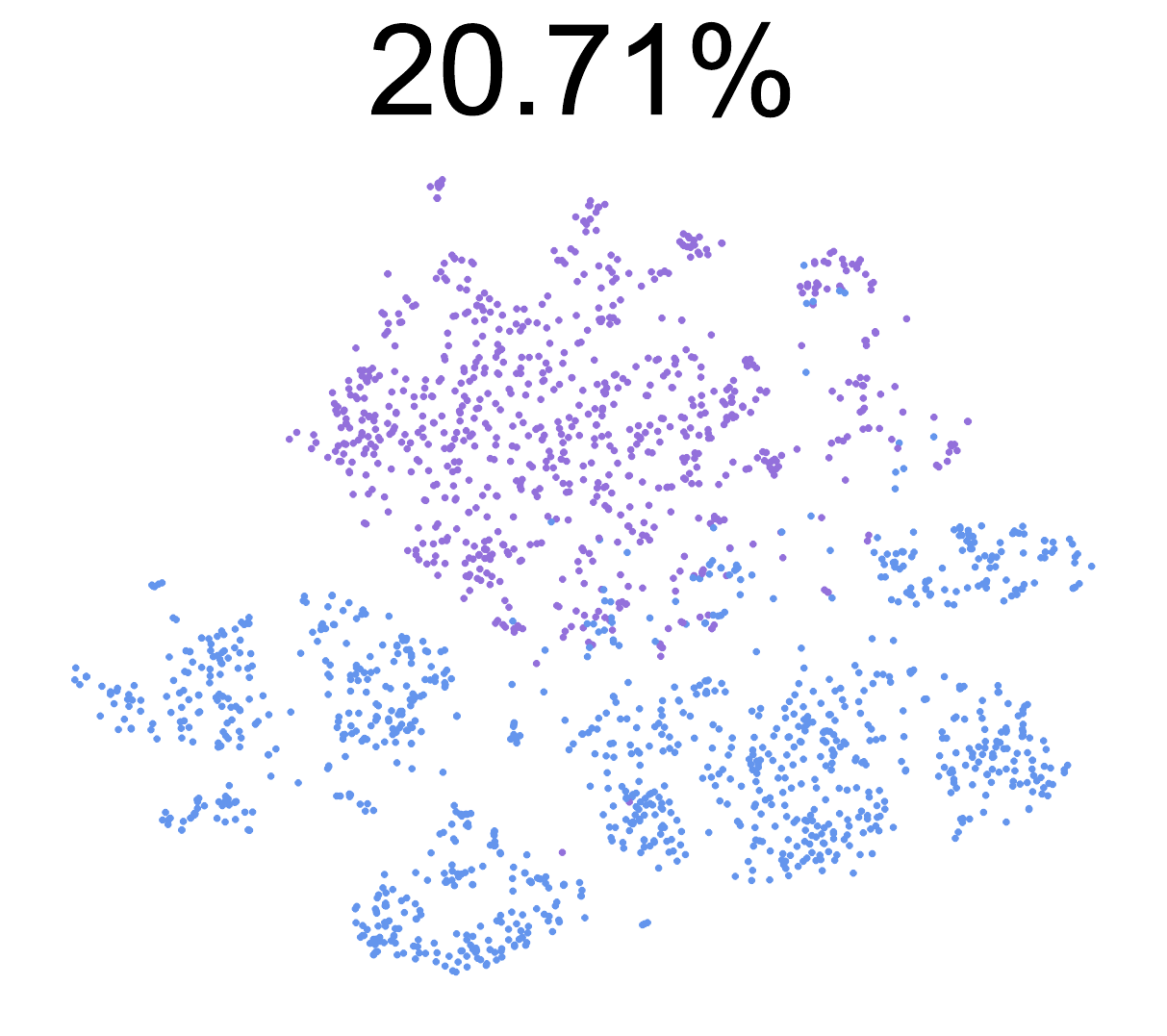}
}
\subfloat[\small{K $=$ 4}]{
\includegraphics[width=0.225\linewidth]{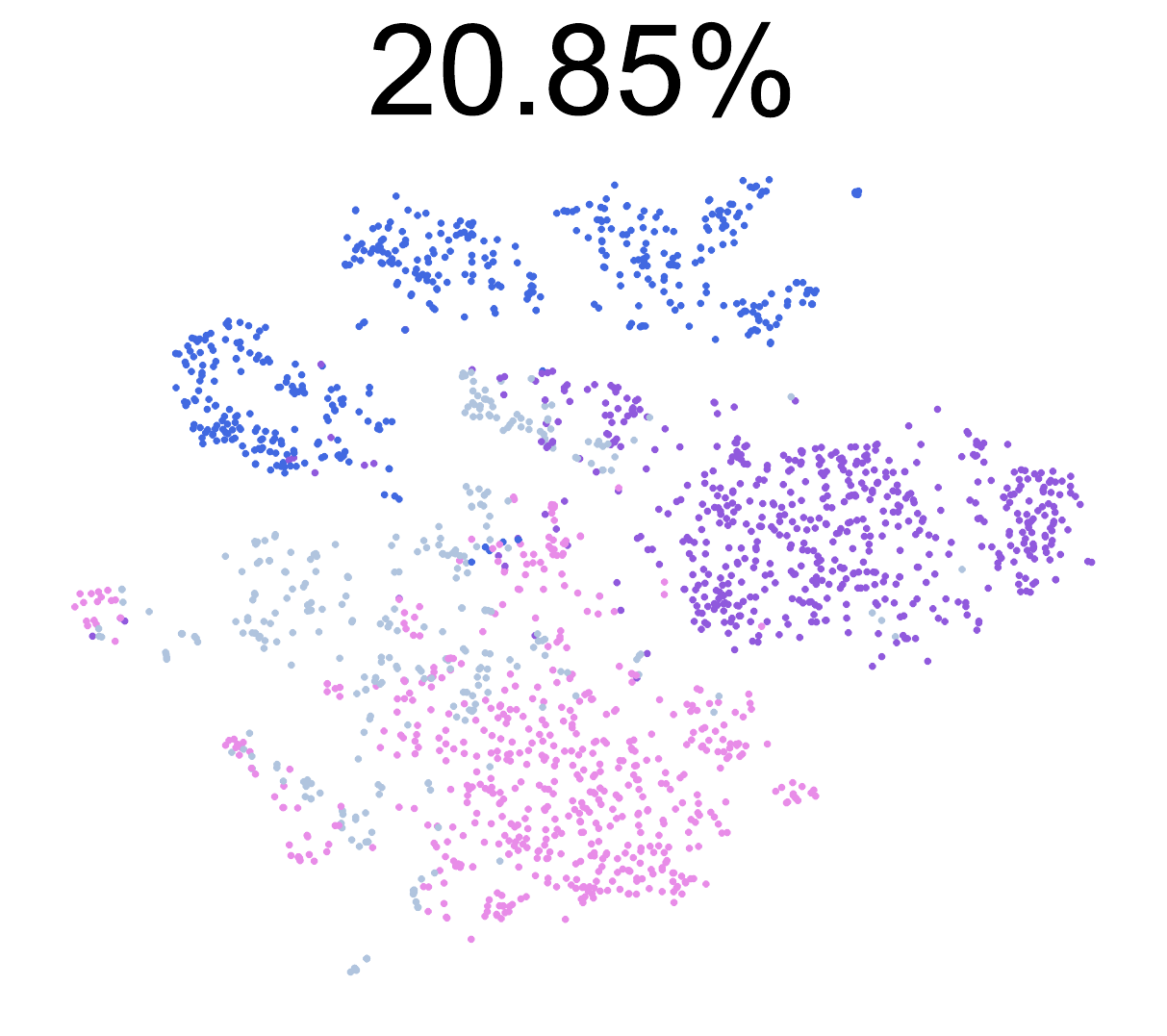}
}
\subfloat[\small{K $=$ 8}]{
\includegraphics[width=0.225\linewidth]{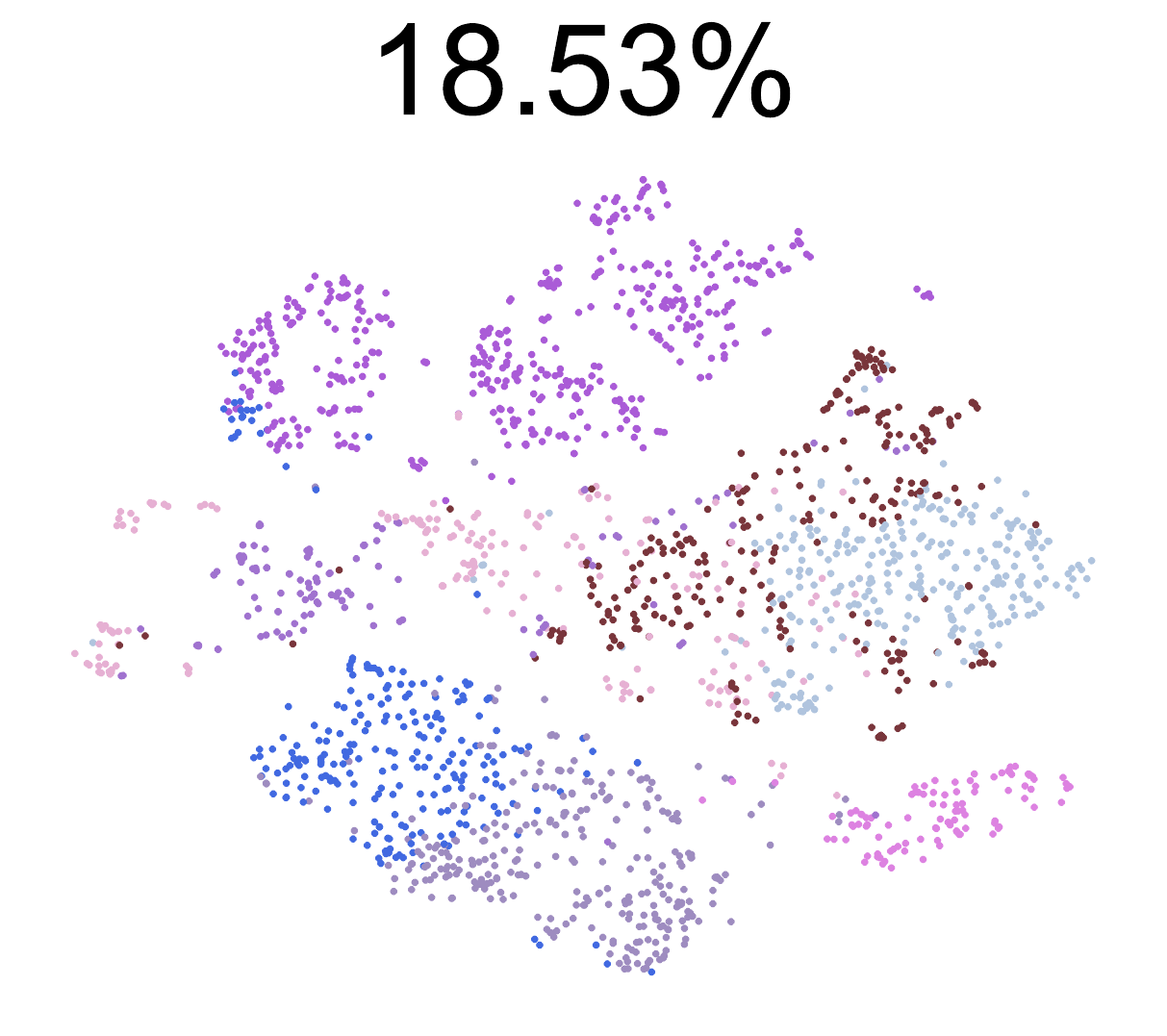}
}
\subfloat[\small{K $=$ 16}]{
\includegraphics[width=0.225\linewidth]{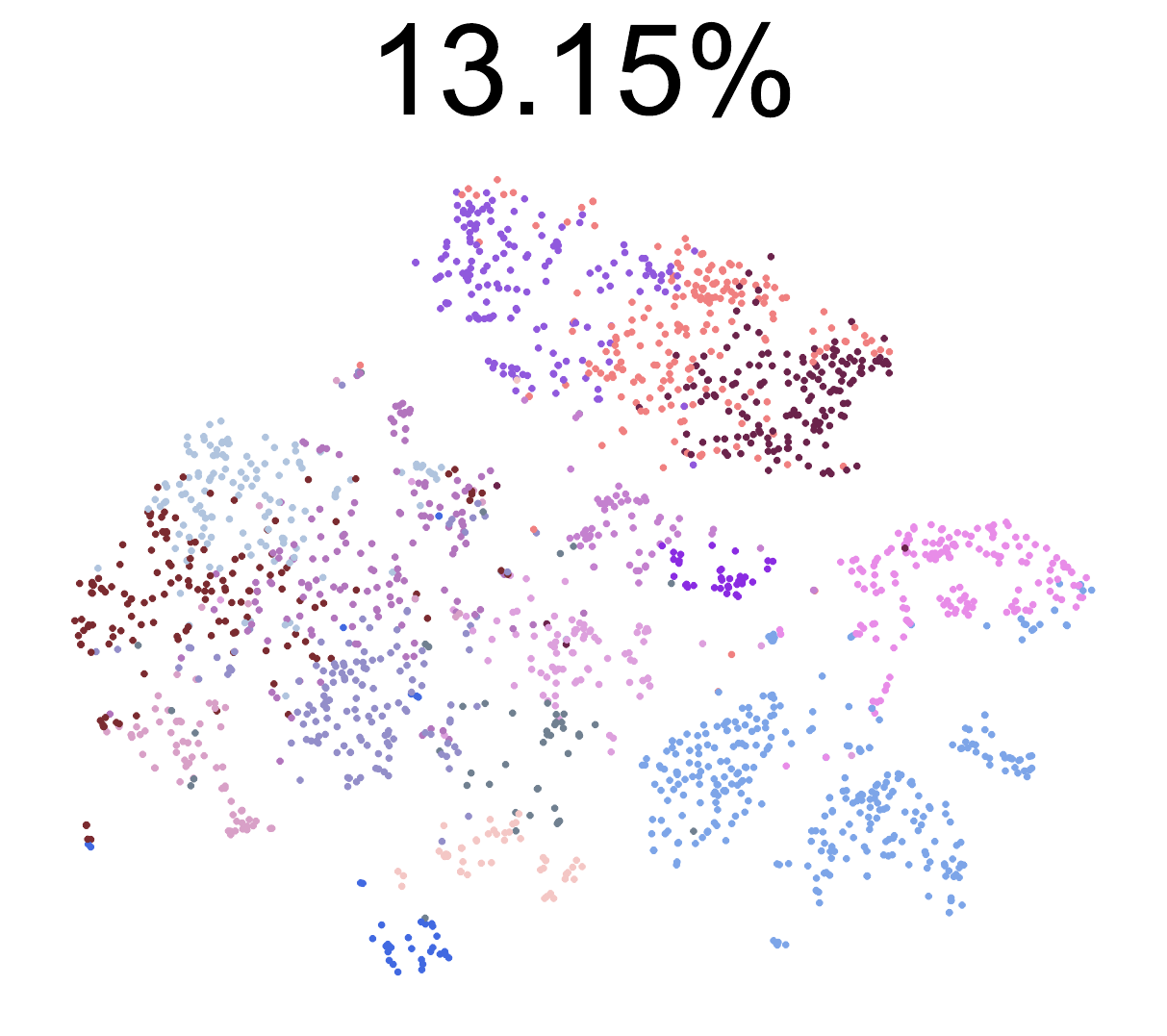}
}

\subfloat[\small{K $=$ 2}]{
\includegraphics[width=0.225\linewidth]{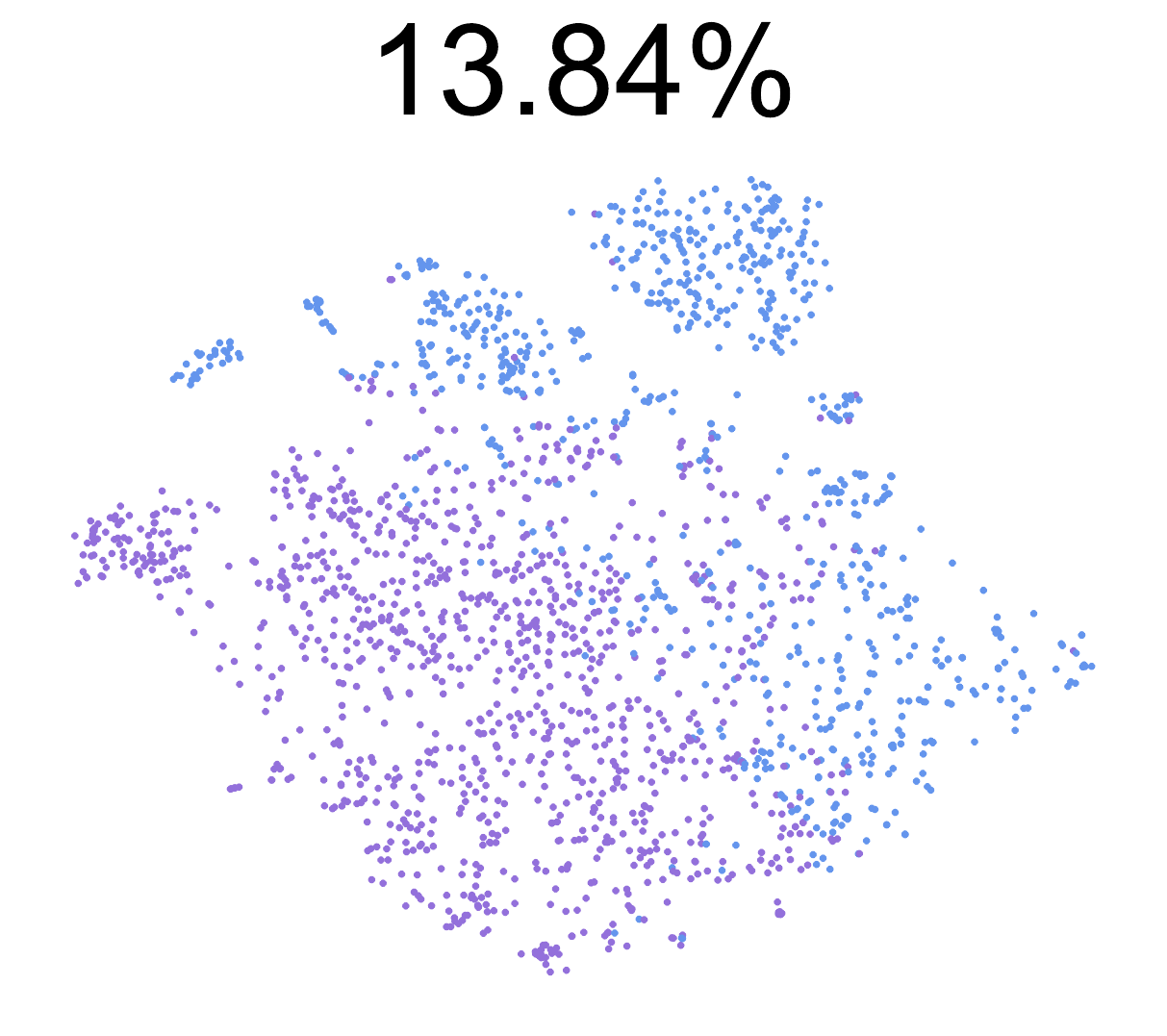}
}
\subfloat[\small{K $=$ 4}]{
\includegraphics[width=0.225\linewidth]{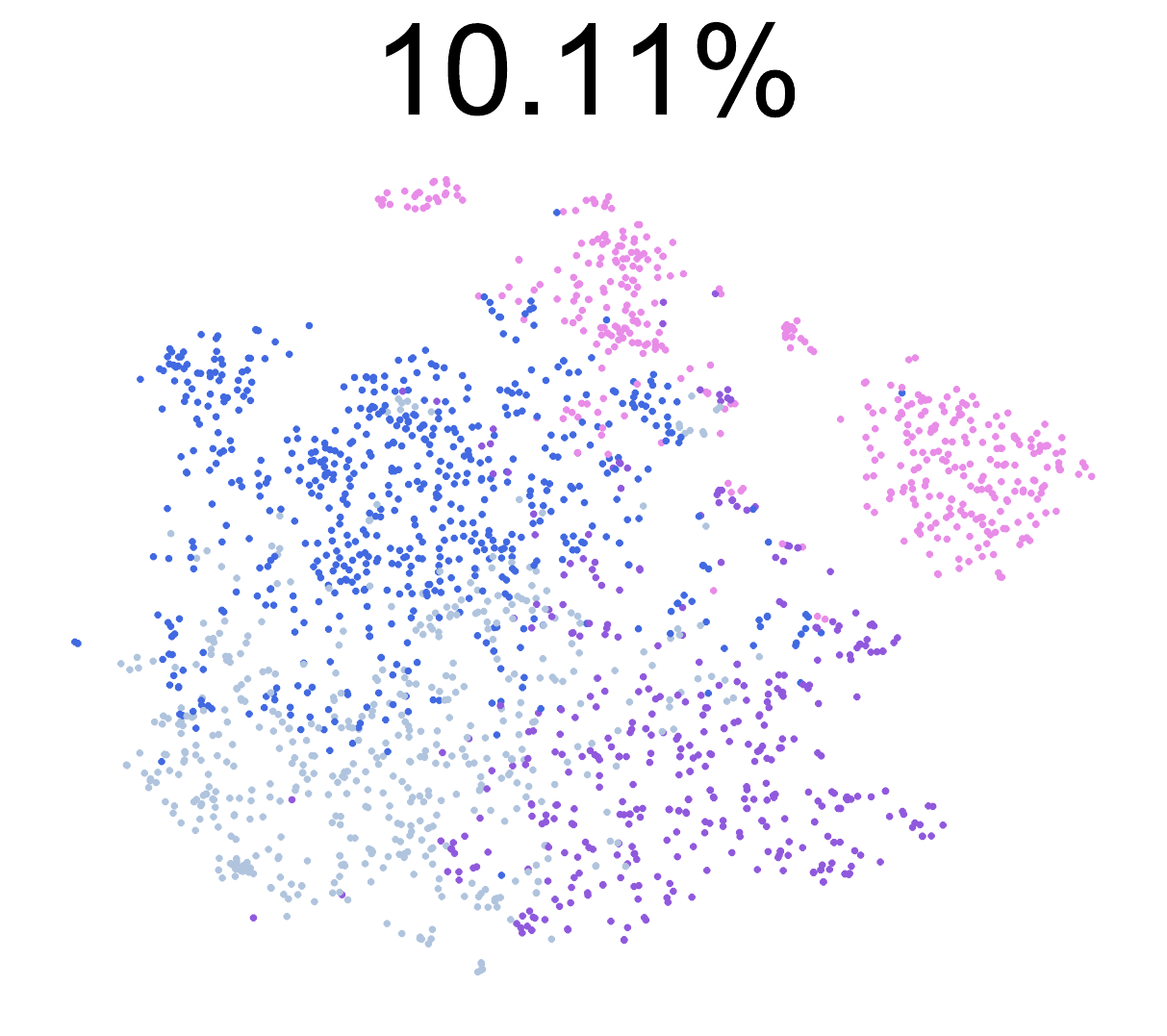}
}
\subfloat[\small{K $=$ 8}]{
\includegraphics[width=0.225\linewidth]{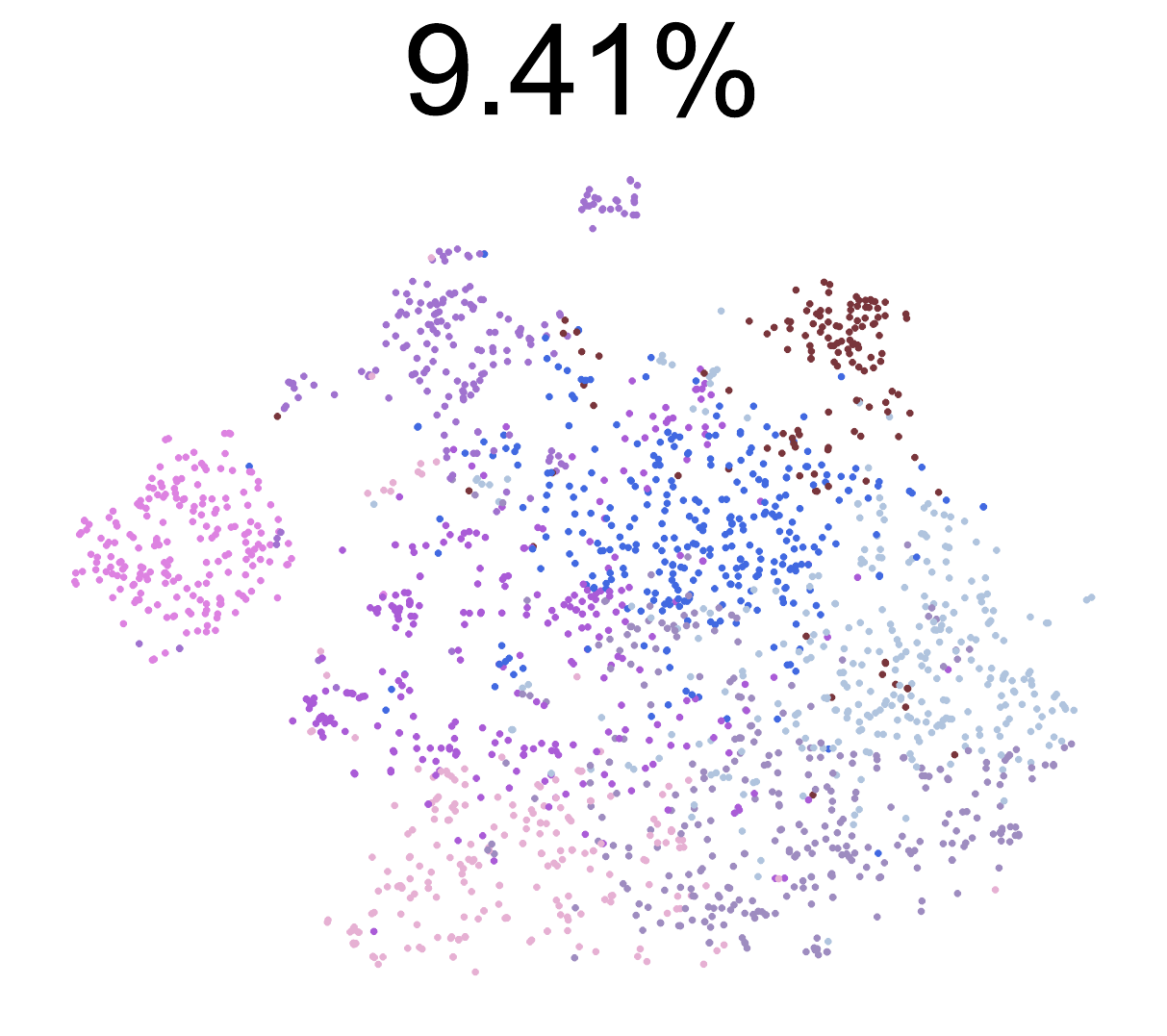}
}
\subfloat[\small{K $=$ 16}]{
\includegraphics[width=0.225\linewidth]{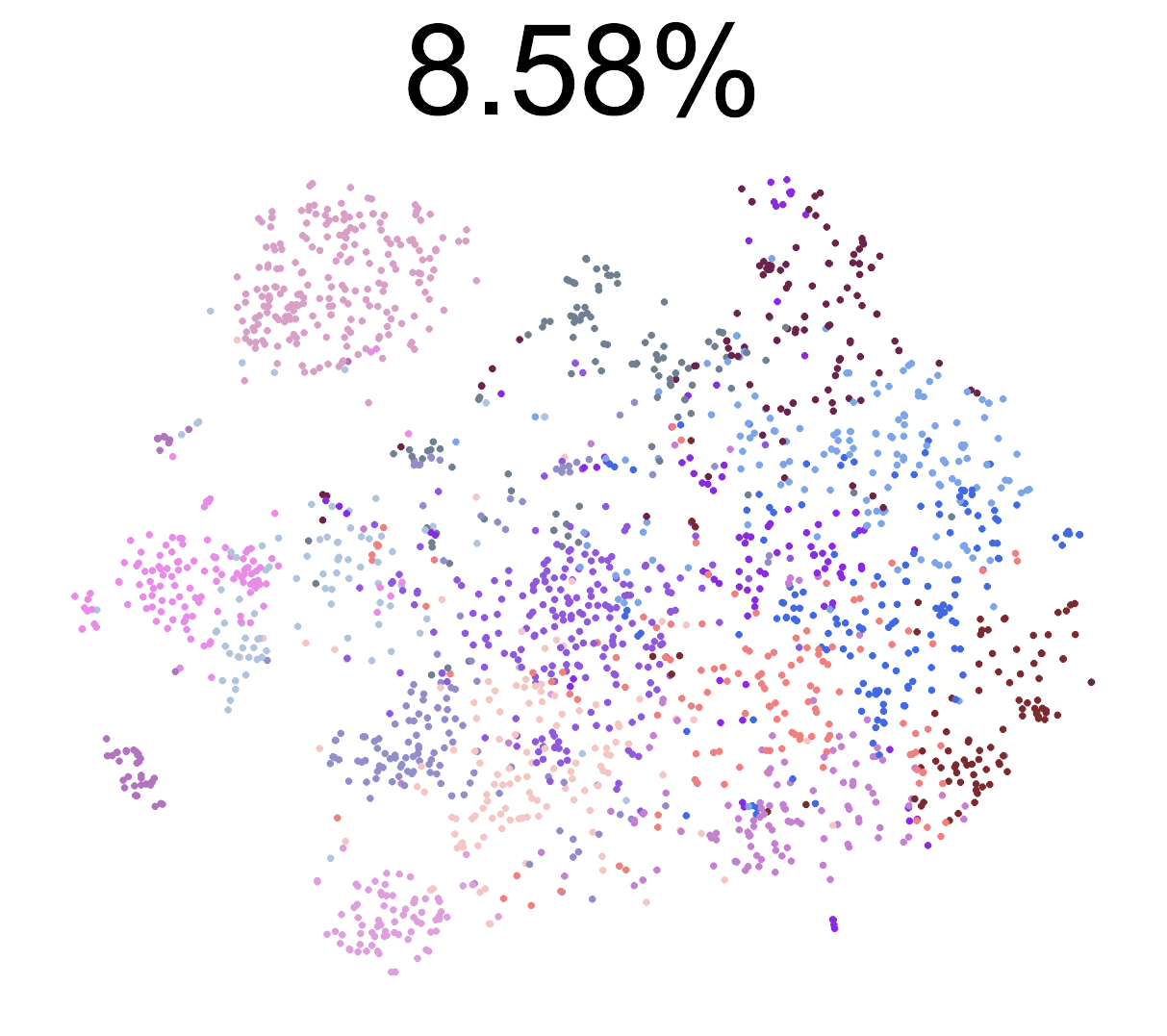}
}
\caption{t-SNE visualization of the reference summary clusters from the train set of
        DISCHARGE (subfigs a--d), 
        ECHO (subfigs e--h) and
        RADIOLOGY (subfigs i--l) 
        with different K-means settings. 
        Silhouette scores are given on the top of each cluster.}
\label{fig:tsne-dis}
\end{figure}

Inspired by GSUM\cite{gsum-model} and Re$^3$Writer\cite{re3writer-model},
we want to leverage sampled summary guidances and enhance the representation learning of the medical report summarization in this study.
Since MIMIC-III is a single-center database, the corresponding medical report and summary may share characteristics of medical writing style or hosiptal report writing paradigm.
To vaild the above hypothesis, 
we first get the contextual representation of the reference summary through pre-trained language model BERT\cite{bert-model}.
Denotes $d$ as the hidden dimension of BERT model,
$S_{train} = \{ T_1...T_N \}$ as the summary set with a total of $N$ reference summaries from the train set of the dataset.
For a reference sequence $T_i$ with length $L$, $T_i = \{t_1 ... t_L \}$, 
its final hidden state from BERT is $y_{i} \in \mathbb{R}^{L \times d}$.
Next, take the text embeddings of the first token, $ y_{i, t_{1}} \in \mathbb{R}^{d}$ as the semantic features of summary $T_i$ because $t_1$ is always a special token \texttt{[CLS]} designated by BERT which represents the beginning of the sequence and attends all the contextual representation learning in the language model.
Consequently, a summary set with selected features $S_{train f} = \{ y_{1, t_{1}} ... y_{N, t_{1}} \}$ is available.
Further, we can estimate the number of clusters in $S_{train}$ with gap statistic\cite{gap-statistic}, 
Table~\ref{tab:gap_value} shows the gap values of gap statistic evaluation, indicating that there could be only one major group.
We visualize the K-means\cite{k-means} clustering results with t-SNE\cite{tsne-visualize} and report the corresponding Silhouette score\cite{silhouette-score} in Fig.~\ref{fig:tsne-dis}.
The overall Silhouette scores remain low, which also supports the hypothesis.

\begin{figure}[h]
    \centering
    \includegraphics[width=0.90\linewidth]{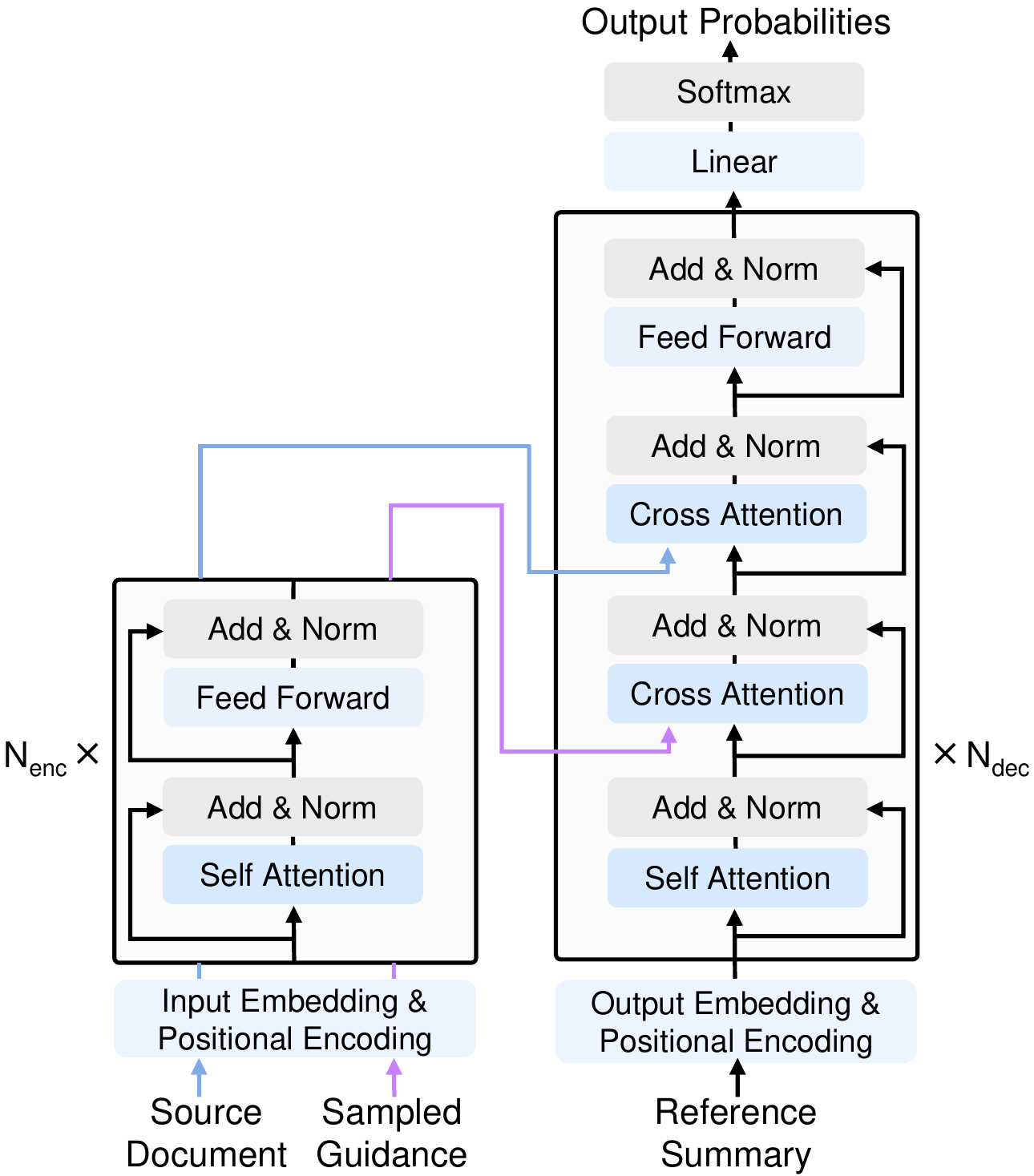}
    \caption{General framework of the transformer-based model with sampled guidance. The encoder (shared parameters) encodes the source document and sampled guidance respectively, while the decoder layer includes additional cross-attention block for the guidance.}
    \label{fig:modelframwork}
\end{figure}

Since the differences of hidden topical semantics from the reference summaries can be indistinguishable under a unsupervised condition,
we can improve the representation learning of the language model with prior summary guidance (i.e., reference summaries from the train set).
Concretely, we consider a randomly sampled reference summary or oracle summary (upper bound of sentence-level extractive summary based on the reference) in $S_{train}$ as the summary guidance for both the training and inference stages.
Based on transformer-based pre-trained language model BART\cite{bart-model}, 
we fine-tune the language model with a simple framework shown in Fig.~\ref{fig:modelframwork}.
Our framework has two main differents to the original transformer-based model:
(i) randomly sample a guidance and encode the guidance with the same encoder for encoding source document;
(ii) let the encoded guidance embeddings do additional cross-attention in the decoder layer of the transformer-decoder.
Thereby no external storage and retrieval module for selecting top-K related guidances is required in our proposed method.
During training, assume $d$ is the hidden dimension of the language model,
denotes $H_{dec} \in \mathbb{R}^{m \times d}$ as the self-attention\cite{transformer-model} output of the hidden representation of output sequence with length $m$ in the decoder layer,
and $H_{g} \in \mathbb{R}^{n \times d}$ as the encoded representation of a guidance with length $n$.
Further, prepare cross-attention\cite{transformer-model} inputs between $H_{dec}$ and $H_{g}$
by linearly projecting $H_{dec}$ into a hidden state $Q = H_{dec} \cdot W^{Q}$, $Q \in \mathbb{R}^{m \times d}$, 
while linearly projecting $H_{g}$ into hiddent states $K = H_{g} \cdot W^{K}$, $K \in \mathbb{R}^{n \times d}$, 
and $V = H_{g} \cdot W^{V}$, $V \in \mathbb{R}^{n \times d}$, respectively.
Computes the attention:
\begin{equation}
\mathit{Attention}(Q, K, V) = \mathit{Softmax}(\frac{Q K^{T}}{\sqrt{d}}) V
\label{eq:cross-attn}
\end{equation}
where the attention output $H_{dec}^{'} \in \mathbb{R}^{m \times d}$ will do another cross-attention with the encoded representation of the source document after residual connection and layer normalization as BART originally designed.
The final optimization objective is the typical cross-entropy loss.

Overall, the proposed simple framework utilizes the prior knowledge of the medical report summarization dataset 
and injects semantic representations of a randomly sampled summary from the train set for both the training and inference stages.

\begin{table}[t]
\centering
\renewcommand\arraystretch{1.1}
\scalebox{0.83}{
    \begin{tblr}
    {colspec=cccccc}
    \hline
    Dataset & K$=$1 & K$=$2 & K$=$4 & K$=$8 & K$=$16
    \\
    \hline
    DISCHARGE & \textbf{1.6395} & 1.7437 & 1.8904 & 2.0331 & 2.1376
    \\
    ECHO & \textbf{1.6157} & 1.8074 & 2.0357 & 2.1580 & 2.3146
    \\
    RADIOLOGY & \textbf{1.5678} & 1.7041 & 1.8190 & 1.9466 & 2.0368
    \\ \hline \\
    \end{tblr}
}
\caption{Gap values of Gap Statistic Evaluation on the reference summary of DISCHARGE, ECHO and RADIOLOGY's train set over different number of clusters (K). 
        The best result is in bold face.
}
\label{tab:gap_value}
\end{table}

\section{Experiments}

\subsection{Implementation Details}
We implement extensive experiments on pre-trained models (BERT2BERT, T5-large, BART) with PyTorch framework.
For BART and T5, 
following their hyperparameter configurations, 
we set the maximum source length of 1024 and the maximum target length of 128 after tokenization.
While BERT2BERT is applied with a maximum source length of 512 and a maximum target length of 128 since the learned positional embeddings of BERT have a maximum length of 512.
We train the models with batch size of 8, learning rate of $5 \times 10^{-5}$, weight decay of 0.01, and AdamW\cite{adamW} optimizer with default settings ($\beta_{1}$ = 0.1, $\beta_{2}$ = 0.999, $\epsilon$ = $10^{-8}$).
We warm up the learning rate with the first 1000 steps (from 0 to $5 \times 10^{-5}$) and then linearly decay afterwards.
For DISCHARGE and RADIOLOGY datasets, we fine-tune the models with 10 epochs,
The ECHO dataset shares the same settings as above, except that it is fine-tuned with 20 epochs.
We validate the model with the evaluation set about every 0.3 epoch, then load and report the test set results of the model checkpoint with the highest ROUGE-1 score.
The models autoregressively generate summaries with beam search (beam width of 6, maximum generation length of 128) for all the datasets.
The experiments are conducted on a single NVIDIA RTX 3090 GPU with mixed precision.

\subsection{Evaluation Metrics}
The summarization outputs are evaluated with the following metrics:
\begin{itemize}
    \item[1)] \textbf{ROUGE} \cite{rouge-metric} is a n-gram-based metric that measures the overlapping n-gram units between the reference summary and candidate summary.
                By computing with the \textit{rouge-score}\footnote{\url{https://pypi.org/project/rouge-score}} package, 
                we report the F1-scores of ROUGE-1 (unigram), ROUGE-2 (bi-gram) and ROUGE-L (the longest common sequence) for the experiments.
    \item[2)] \textbf{BERTScore} \cite{bertscore-metric} is a semantic similarity metric based on the pre-trained language model BERT that evaluates the similarity between the contextual embeddings of reference text and machine-generated text.
                We use \textit{bert-score}\footnote{\url{https://pypi.org/project/bert-score}} package for the computation.
    \item[3)] \textbf{SummaC} \cite{summac-metric} is a reference-free consistency evaluation method that detects inconsistencies between the source document and machine-generated summary with a pre-trained natural language inference (NLI) model.
                The method constructs a NLI pair matrix across every sentence from the source document and machine-generated summary, 
                reflecting the properties (entailment, contradiction and neutrality) of these sentnece pairs.
                We use SummaC\textsubscript{CONV} (abbreviated as SummaC) for the experiments, it fowards the NLI pair matrix to a learned 1-D convolutional layer and calculates the average consistency score of the candidate summary\footnote{\url{https://github.com/tingofurro/summac}}.
    \item[4)] \textbf{QuestEval} \cite{questeval-metric} is a reference-free summarization metric for evaluating factual consistency. 
                The method consists of a pre-trained question-generation (QG) model and a pre-trained question-answering (QA) model:
                the QG model creates correlated questions from the candidate summary, 
                and then the QA model generates corresponding answers. 
                QuestEval examines to what extent the questions from a candidate summary can be answered by its source,
                taking into account the named entities and nouns from the source document as the ground-truth answers\footnote{\url{https://github.com/ThomasScialom/QuestEval}}.
\end{itemize}

\subsection{Compared Methods and Parameters}
We take BART as the encoder-decoder baseline and add a cross-attention block in each of its decoder layers for the proposed summary guidance on language model fine-tuning framework.
We abbreviate the method as BART\scriptsize{$+$Oracle Guidance}\normalsize{} when the guidance is oracle summary,
and BART\scriptsize{$+$Ref Guidance}\normalsize{} when the guidance are reference summary. Note that all the summary guidances are randomly sampled from the train set.

We compare the proposed framework with the following summarization methods:
\begin{itemize}
    \item[1)] \textbf{Lead-3}: a simple extractive summarization baseline that takes the leading three sentences from the source document as the summary.
    \item[2)] \textbf{Oracle}: the upper bound of extractive summarization. Oracle summary can be obtained with greedy search that considers maximizing ROUGE scores as the objective when the human-written summary is given.
    \item[3)] \textbf{BERT2BERT}: an encoder-decoder model that takes pre-trained BERT-base\cite{bert-model} model as both the encoder and the decoder.
                Pre-trained on BooksCorpus and English Wikipedia, BERT-base (110M parameters) is an encoder model with 12 layers, hidden state size of 768 and a vocabulary size of 30K, 
                hence BERT2BERT has 220M of parameters.
    \item[4)] \textbf{T5-large}\cite{t5-model}: a transformer-based encoder-decoder model pre-trained on large clean corpus with masked language modeling, the model casts multiple NLP training tasks info a Seq2Seq paradigm.
                T5-large (770M parameters) has 24 layers of encoder, 24 layers of decoder, hidden state size of 1024 and a vocabulary size of 32K.
    \item[5)] \textbf{BART}\cite{bart-model}: a self-supervised transformer-based encoder-decoder model pre-trained on books, news, stories and web text with masked language modeling.
                BART (406M parameters) has 12 layers of encoder, 12 layers of decoder, hidden state size of 1024 and a vocabulary size of 50K.
\end{itemize}


\subsection{Experimental Results}

\newsavebox\CBox
\def\textBF#1{\sbox\CBox{#1}\resizebox{\wd\CBox}{\ht\CBox}{\textbf{#1}}}

\begin{figure*}[!htbp]
    \begin{minipage}[t][470pt]{1.0\textwidth}
    \centering
    \begin{table}[H]
    \small
    \centering
    \subfloat[\small{DISCHARGE}\label{result on dis}]{
    \scalebox{1.00}{
    \begin{tblr}{
        colspec={lccccccr}, 
        row{9,10} = {bg=gray9},
        rowsep=0.5pt, 
        stretch=0, 
        rows={ht=\baselineskip}
    }
    \hline
    Model & R-1 & R-2 & R-L & BERTScore & SummaC & QuestEval & Training
    \\
    \hline
    \SetCell[c=8]{l}$Extractive$
    \\ \hline
    Lead-3 & \, 8.85 & \, 3.26 & \, 8.65 & 79.77 & \textbf{94.78} & \textbf{37.15} & - \\
    Oracle & 28.86 & 15.64 & 27.42 & 82.35 & 84.89 & 23.66 & - \\
    \hline
    \SetCell[c=8]{l}$Abstractive$
    \\ \hline
    BERT2BERT & 26.80 & 12.59 & 25.12 & 82.96 & 52.47 & 20.15 & \textBF{22 hrs} \\
    T5-large & 35.76 & \textbf{20.89} & 33.72 & 84.55 & 63.48 & 21.60 & 64 hrs \\
    BART & 35.62 & 19.96 & 35.23 & 85.83 & 60.52 & 21.21 & 24 hrs \\
    BART\scriptsize{$+$Oracle Guidance} & 36.10 & 20.06 & 35.69 & \textbf{85.90} & 61.61 & 21.26 & 32 hrs \\
    BART\scriptsize{$+$Ref Guidance} & \textbf{36.21} & 20.27 & \textbf{35.88} & 85.84 & 58.16 & 21.26 & 32 hrs \\
    \hline
    $Reference$ &  &  &  &  & 55.67 & 20.63 &  
    \\ \hline
    \end{tblr}}}
  
    \subfloat[\small{ECHO}\label{result on echo}]{
    \scalebox{1.00}{
    \begin{tblr}{
        colspec={lccccccr}, 
        row{9,10} = {bg=gray9},
        rowsep=0.5pt, 
        stretch=0, 
        rows={ht=\baselineskip}
    }
    \hline
    Model & R-1 & R-2 & R-L & BERTScore & SummaC & QuestEval & Training
    \\
    \hline
    \SetCell[c=8]{l}$Extractive$
    \\ \hline
    Lead-3 & 16.45 & \, 3.98  & 15.49 & 82.68 & \textbf{96.08} & \textbf{42.04} & - \\
    Oracle & 46.45 & 23.44 & 40.77 & 85.74 & 85.61 & 33.07 & - \\
    \hline
    \SetCell[c=8]{l}$Abstractive$
    \\ \hline
    BERT2BERT & 53.31 & 39.20 & 52.15 & 88.85 & 59.53 & 23.51 & \textBF{7 hrs} \\
    T5-large & 55.63 & \textbf{42.05} & 54.45 & 89.73 & 63.94 & 24.02 & 36 hrs \\
    BART & 55.22 & 39.71 & 54.50 & 90.15 & 60.24 & 23.70 & \textBF{7 hrs} \\
    BART\scriptsize{$+$Oracle Guidance} & \textbf{55.74} & 40.23 & 54.95 & 90.17 & 59.53 & 23.94 & 11 hrs \\
    BART\scriptsize{$+$Ref Guidance} & 55.71 & 40.45 & \textbf{54.97} & \textbf{90.21} & 60.00 & 23.76 & 11 hrs \\
    \hline
    $Reference$ &  &  &  &  & 57.04 & 23.52 &  
    \\ \hline
    \end{tblr}}}

    \subfloat[\small{RADIOLOGY}\label{result on rad}]{
    \scalebox{1.00}{
    \begin{tblr}{
        colspec={lccccccr}, 
        row{9,10} = {bg=gray9},
        rowsep=0.5pt, 
        stretch=0, 
        rows={ht=\baselineskip}
    }
    \hline
    Model & R-1 & R-2 & R-L & BERTScore & SummaC & QuestEval & Training
    \\
    \hline
    \SetCell[c=8]{l}$Extractive$
    \\ \hline
    Lead-3 & 15.36 & \, 4.78 & 14.55 & 83.46 & \textbf{88.96} & \textbf{37.59} & - \\
    Oracle & 40.12 & 22.07 & 35.83 & 85.94 & 81.55 & 34.19 & - \\
    \hline
    \SetCell[c=8]{l}$Abstractive$
    \\ \hline
    BERT2BERT & 45.37 & 28.01 & 43.28 & 87.23 & 38.61 & 24.36 & \textBF{38 hrs} \\
    T5-large & \textbf{51.51} & \textbf{32.40} & 49.28 & 88.75 & 40.47 & 25.51 & 251 hrs \\
    BART & 51.34 & 31.59 & 49.88 & 89.91 & 37.92 & 25.20 & 55 hrs \\
    BART\scriptsize{$+$Oracle Guidance} & 51.46 & 31.71 & \textbf{49.99} & \textbf{89.94} & 37.88 & 25.20 & 73 hrs \\
    BART\scriptsize{$+$Ref Guidance} & 51.41 & 31.71 & 49.95 & \textbf{89.94} & 37.94 & 25.26 & 73 hrs \\
    \hline
    $Reference$ &  &  &  &  & 32.25 & 23.58 &  
    \\ \hline
    \end{tblr}}}
    \vspace{0.5em}
    \caption{
      The results of ROUGE scores, BERTScore and SummaC and QuestEval on the test set of DISCHARGE, ECHO and RADIOLOGY datasets.
      Time consumption of the training stage is given in the last column.
      The best result is in bold face.
    }
    \label{tab:main result}
    \end{table}
    \end{minipage}
\end{figure*}

In Table~\ref{tab:main result}, we present the evaluation results on extractive summarization and abstractive summarization,
Since both SummaC and QuestEval are reference-free factual consistency metrics (i.e., evaluating coherence between source document and candidate summary),
we also report their evaluation outcome on the ground-truth summary.

The result of ROUGE shows that the overall performance of abstractive models is significantly better than extractive methods Lead-3 and Oracle, except BERT2BERT on DISCHARGE (Table~\ref{result on dis}), 
this may be because BERT2BERT has relatively small vocabulary size in the compared models and fewer learned positional embeddings.
Second, T5 holds the top R-2 score over three datasets, this can be due to the model has double the number of transformer layers than BART.
Third, our proposed method can increase R-1 and R-L on BART for about 0.4 to 0.6 on both DISCHARGE (Table~\ref{result on dis}) and ECHO (Table~\ref{result on echo}), surpassing T5-large's.
Next, the proposed method creates a marginal increase in ROUGE scores on RADIOLOGY (Table~\ref{result on rad}), which could be because the language model reaches its bottleneck, since T5-large also does not present a great lead over BART.
Furthermore, BART\scriptsize{$+$Oracle Guidance}\normalsize{} and BART\scriptsize{$+$Ref Guidance}\normalsize{} can develop similar level of improvement on R-1, 
but about 0.2 higher increases on R-2 for DISCHARGE and ECHO datasets are achieved with BART\scriptsize{$+$Ref Guidance}\normalsize{}.

The result of BERTScore is basically consistent with that of ROUGE, 
and the proposed BART\scriptsize{$+$Oracle Guidance}\normalsize{} and BART\scriptsize{$+$Ref Guidance}\normalsize{} can produce summaries that are closer to the human-written criteria.
Moreover, it can be found that BERTScore has a narrower range of results across various summarization methods than ROUGE's,
making it difficult to reason to what extent the models' performance differ.
Although T5-large can offer a higher R-2 result, 
the corresponding contextual and semantic similarity to the ground truth summary is generally lower than the original BART.

The results of reference-free metrics SummaC and QuestEval might deviate from ROUGE and BERTScore.
First, the reference summaries have a low factual consistency score to the source document, 
this could be due to SummaC and QuestEval failing to generalize on the medical corpus,
and the corresponding inconsistency detection may be sensitive to medical proper nouns that share in few sentences hence the metrics consider the other lines could be ``unfamiliar'' and ``conflict'' with the proper nouns.
SummaC and QuestEval would predict the veracity by evaluating whether the claims (generally determined by nouns and verbs) from the candidate summary could be supported by the source document at a sentence-level,
accordingly the reference summary with well-chosen medical terms could have lower factual consistency scores than machine-generated summaries in this study.
The following section~\ref{sec:case study} shows cases of the overlapping circumstance between the source and the reference summary, 
as well as typical appearances of proper nouns across the source document.
Second, Lead-3 and Oracle achieve high-level factual consistency by merely copying a few original lines from the source as the summary rather than compressing and paraphrasing the context into a free-form summary, 
which is unlikely to conflict with the source material.
Further, T5-large achieves overall better factual consistency for the source document,
meanwhile the proposed method maintains a level of factual consistency comparable to that of the original BART.

Overall, the comprehensive experimental results suggest that the proposed sampled guidance on language model fine-tuning can gain decent ROUGE scores improvement and maintain the original level of factual consistency with an increase of about 35\% running time, 
boosting the baseline model BART surpassing T5-large.
An alternative way to efficiently leverage the representation of sampled summary guidance is storing the encoder output of summary guidances locally and then sampling the encoded embeddings at training and inference stage,
but can also considerably increase the storage occupancy.
Besides, BERT2BERT is a time-efficient model with fair performance for ECHO and RADIOLOGY, but a pre-trained model with longer maximum positional embeddings would be a favorable choice for DISCHARGE.
Although the T5-large has the highest R-2 and greater factual consistency results over the abstractive methods,
a much longer training time consumption should not be disregarded.

\subsection{Case Study}
\label{sec:case study}
In this section, we demonstrate a few generated examples on the test set of DISCHARGE (Table~\ref{tab:case-dis}), ECHO (Table~\ref{tab:case-echo}) and RADIOLOGY (Table~\ref{tab:case-rad}).
The index of the example is indicated by the number after the symbol \#.
The overlapping contexts are highlighted in light red, blue and green.

\begin{table}[h]
\centering
\renewcommand\arraystretch{1.00}
\scalebox{0.92}{
\begin{tabular}{p{0.5\textwidth}}
\hline
\textbf{Source:}
Service: cardiac surgery
history of present illness: [...] 
an angiogram after the stent showed an aortic valve disease.
six months ago she was hospitalized with \textcolor{LightCoral}{congestive heart failure} and treated with lasix. 
[...]
Hospital Course: the patient was taken to the
operating room where she had an \textcolor{DodgerBlue3}{aortic valve replacement
surgery with porcine valve}. the indications for surgery were
an \textcolor{DodgerBlue3}{aortic stenosis} with valve area less than .5 and chf and
symptomatic severe aortic stenosis with shortness of breath
at rest, edema and occasionally cough. 
[...] 
Medications: [...]
\textcolor{ForestGreen}{she was being treated for e. coli in her uti and sputum h flu }
and found to have enterococcus and we added ampicillin to her
antibiotic regimen.
[...]
Discharge Instructions:
[...]
\textcolor{ForestGreen}{she should get a repeat urinalysis} and
she should continue cipro for 9 additional days.
[...]
\\ \hline
\textbf{Reference:}
1.  \textcolor{DodgerBlue3}{status post avr with hancok porcine valve}.
2.  \textcolor{ForestGreen}{urinary tract infection}.
3.  \textcolor{LightCoral}{congestive heart failure}.
\\ \hline
\textbf{BERT2BERT:}
1.  aortic stenosis.
2.  coronary artery disease.
3.  \textcolor{LightCoral}{congestive heart failure}.
4.  type 2 diabetes.
5.  hypertension.
6.  high cholesterol.
\\ \hline
\textbf{T5-large:}
aortic stenosis \textcolor{DodgerBlue3}{status post aortic valve} replacement with \textcolor{DodgerBlue3}{porcine valve}.
\\ \hline
\textbf{BART:} 
1. aortic stenosis.
2. \textcolor{DodgerBlue3}{status post} aortic valve replacement.
3. postoperative atrial fibrillation.
4. postoperative leukocytosis.
\\ \hline
\textbf{BART\scriptsize{$+$Oracle Guidance}:}
1.  aortic stenosis \textcolor{DodgerBlue3}{status post porcine avr}.
2.  \textcolor{ForestGreen}{urinary tract infection}.
condition on discharge:  good.
\\ \hline
\textbf{BART\scriptsize{$+$Ref Guidance}:}
1.  aortic stenosis \textcolor{DodgerBlue3}{status post porcine avr}.
2.  type 2 diabetes.
3.  hypertension.
4.  high cholesterol.
5.  coronary artery disease.
6.  \textcolor{LightCoral}{congestive heart failure}.
7.  \textcolor{ForestGreen}{urinary tract infection}.
\\ \hline \\
\end{tabular}
}
\caption{Generated examples of DISCHARGE. \#102.}
\label{tab:case-dis}
\end{table}

\begin{table}[h]
\centering
\renewcommand\arraystretch{1.00}
\scalebox{0.92}{
\begin{tabular}{p{0.5\textwidth}}
\hline
\textbf{Source:}
Findings:
right atrium/interatrial septum: no asd by 2d or color doppler.
left ventricle: overall normal lvef (>55\%).
Sorta: no atheroma in descending aorta.
Aortic Valve: mildly thickened aortic valve leaflets (3). 
\textcolor{DodgerBlue3}{no masses or vegetations on aortic valve}. no ar.
Mitral Valve: normal mitral valve leaflets. 
\textcolor{DodgerBlue3}{no mass or vegetation on mitral valve}. mild (1+) mr.
Tricuspid Valve: normal tricuspid valve leaflets.
[...]
Conclusions: [...] 
\textcolor{ForestGreen}{no aortic regurgitation is seen}. 
the mitral valve leaflets are structurally normal. 
no mass or vegetation is seen on the mitral valve. 
\textcolor{ForestGreen}{mild (1+) mitral regurgitation is seen}. 
there is no pericardial effusion.
\\ \hline
\textbf{Reference:}
\textcolor{DodgerBlue3}{no mass or vegetation visualized}.
\textcolor{ForestGreen}{normal global lv systolic function with mild mitral regurgitation}.
\\ \hline
\textbf{BERT2BERT:}
no evidence of endocarditis.
\textcolor{ForestGreen}{mild mitral regurgitation}.
\\ \hline
\textbf{T5-large:}
no valvular \textcolor{DodgerBlue3}{vegetations} seen.
\textcolor{ForestGreen}{mild mitral regurgitation}.
\\ \hline
\textbf{BART:}
no echocardiographic evidence of endocarditis.
\textcolor{ForestGreen}{mild mitral regurgitation}.
mild tricuspid regurgitation.
\\ \hline
\textbf{BART\scriptsize{$+$Oracle Guidance}:}
no \textcolor{DodgerBlue3}{vegetations} seen.
\textcolor{ForestGreen}{mild mitral regurgitation}.
\\ \hline
\textbf{BART\scriptsize{$+$Ref Guidance}:}
no echocardiographic evidence of endocarditis.
\textcolor{ForestGreen}{mild mitral regurgitation}.
\\ \hline \\
\end{tabular}
}
\caption{Generated examples of ECHO. \#83.}
\label{tab:case-echo}
\end{table}

\begin{table}[h]
\centering
\renewcommand\arraystretch{1.00}
\scalebox{0.92}{
\begin{tabular}{p{0.5\textwidth}}
\hline
\textbf{Source:}
Final Report:
Indication: persistent right upper lobe opacity on chest radiograph with new
presentation of blood tinged sputum. history of hiv.
[...]
Chest CT w/o Contrast: \textcolor{ForestGreen}{there is no axillary, hilar or mediastinal adenopathy}.
there continues to be volume loss of the right lung as seen on the chest ct of
[**2158-5-17**]. there is \textcolor{DodgerBlue3}{slight decrease in size of the opacity in the lateral
segment of the right middle lobe which is associated with adjacent
bronchiectasis}. there is a small focal opacity in the superior segment of the
right lower lobe. \textcolor{ForestGreen}{there are no lung nodules or masses}. there are no
endobronchial lesions. the trachea and segmental bronchi are patent.
\textcolor{LightCoral}{there is scattered ground glass opacity primarily in both upper lobes}. 
heart size is normal. a 4 cm simple cyst is present in the upper pole of the right
kidney. the upper abdominal organs are incompletely visualized.
\\ \hline
\textbf{Reference:}
Impression: 
\textcolor{DodgerBlue3}{slight decrease} since [**2158-5-17**] of the \textcolor{DodgerBlue3}{opacity in the lateral
segment of the right middle lobe}, probably an area of persistent atelectasis.
\textcolor{ForestGreen}{there are no obstructing masses or adenopathy}.
\textcolor{LightCoral}{bilateral ground glass opacity in the upper lobes could represent an infectious process and/or chf}.
\\ \hline
\textbf{BERT2BERT:}
Impression: 
1) there are no lung nodules or masses.
2) there continues to be volume loss in the right lung with a focal area of bronchiectasis \textcolor{DodgerBlue3}{in the lateral
segment of the right middle lobe} and a smaller area in the superior segment of the right lower lobe.
3) there is \textcolor{LightCoral}{scattered ground glass opacity in both upper lobes}.
\\ \hline
\textbf{T5-large:}
Impression: 
1) \textcolor{DodgerBlue3}{slight decrease in size of the opacity in the lateral segment of the right middle lobe} which is associated with adjacent bronchiectasis.
2) small focal opacity in the superior segment of the right lower lobe.
3) \textcolor{LightCoral}{scattered ground glass opacities in both upper lobes}.
this may represent an \textcolor{LightCoral}{infectious} or inflammatory process.
\\ \hline
\textbf{BART:}
Impression:
1) \textcolor{DodgerBlue3}{slight decrease in size of the opacity in the lateral segment of the right middle lobe} which is associated with bronchiectasis.
there are also small focal opacities in the superior segment of the right lower lobe which are nonspecific and may represent an \textcolor{LightCoral}{infectious} or inflammatory process.
2) volume loss of the right lung as seen on the chest ct of [**2158-5-17**].
3) simple cyst of the upper pole of the right kidney.
\\ \hline
\textbf{BART\scriptsize{$+$Oracle Guidance}:}
Impression:
1) \textcolor{DodgerBlue3}{slight decrease in size of the opacity in the lateral segment of the right
middle lobe} which is associated with adjacent bronchiectasis.
this likely represents a slowly resolving \textcolor{LightCoral}{infectious} process.
there is no evidence of endobronchial lesions.
2) there are \textcolor{LightCoral}{scattered ground glass opacities in both upper lobes} which may be
related to the patient's hiv status.
\\ \hline
\textbf{BART\scriptsize{$+$Ref Guidance}:}
Impression:
1) persistent volume loss in the right lung with \textcolor{DodgerBlue3}{slight decrease in size of the
opacity in the lateral segment of the right middle lobe} associated with
bronchiectasis. there is no evidence of endobronchial lesion.
2) \textcolor{LightCoral}{scattered ground glass opacities in both upper lobes} consistent with an \textcolor{LightCoral}{infectious} or inflammatory process.
\\ \hline \\
\end{tabular}
}
\caption{Generated examples of RADIOLOGY. \#51.}
\label{tab:case-rad}
\end{table}

Table~\ref{tab:case-dis} shows that these abstractive summarization model. can recapitulate the critical phrases as bullet points. 
Further, BART\scriptsize{$+$Oracle Guidance}\normalsize{} and BART\scriptsize{$+$Ref Guidance}\normalsize{} can recognize more accurate phrases than the others.
Table~\ref{tab:case-echo} present a case that not every baseline model can identify the key lines even though the target reference summary may be fairly basic and extractive,
since only T5-large and BART\scriptsize{$+$Oracle Guidance}\normalsize{} view ``no vegetations'' as the primary diagnosis.
Table~\ref{tab:case-rad} is a challenging case that BERT2BERT simply copies the line ``there are no lung nodules or masses'' as the first summary sentence.
Although most models successfully spot the sentence ``slight decrease ... in the lateral segment ...'',
they fail to reason the symptoms as ``persistent atelectasis''.
This is a typical issue with language models, they can pay attention to the salient information in the text but cannot reason.

Overall, the above cases indicate that the proposed method can enhance the summarization quality of the original BART model.
Furthermore, the objective of our approach is to optimize the language model with the additional summarization guidance signal from the train set,
which may not directly enhance the language model's ability in knowledge-based inference.
However, the proposed method can be complementary to the fact-aware and knowledge-based summarization approaches,
which is worthy of future in-depth investigation.

\section{Limitation and Conclusion}

This paper leveraged the summary guidance from the train set of the medical report dataset, 
and a domain-specific medical report dataset is unlikely to cover a broad range of writing styles. 
Next, text generation datasets with a variety of topics may require an additional step of topical retrieval or clustering for the proposed method. 
Further, the summary guidance may not cure language models' issue of producing unfaithful arguments.

In conclusion, this study proposed three medical report-summary datasets DISCHARGE, ECHO and RADIOLOGY that were derived from MIMIC-III.
Further, we fine-tuned pre-trained language models BERT2BERT, T5-large, and BART on the datasets to establish strong baselines for medical abstractive summarization. 
In addition, we statistically and intuitively showed that the summaries could share robust characteristics, with their corresponding hidden topical semantics difficult to cluster into multiple groups. 
To further improve the performance, the study proposed a simple framework that uses a randomly selected reference summary from the training set as a summary guidance for both the training and inference stages. 
The extensive experiments demonstrated the effectiveness and feasibility of the proposed method, 
as it showed decent improvement in ROUGE and BERTScore while not observing a significant decrease in faithfulness score.

\nolinenumbers

\vfill

\begin{thebibliography}{1}
\bibliographystyle{IEEEtran}

\bibitem{standford-sum}{
    Y. Zhang, D. Y. Ding, T. Qian, C. D. Manning, and C. P. Langlotz,
    ``Learning to Summarize Radiology Findings,''
    in \textit{Proc. Workshop Health Text Mining Info. Analy.}, 2018, pp. 204--213.
}
\bibitem{multi-doc-sum}{
    H. C. Shing, C. Shivade, N. Pourdamghani, F. Nan, P. Resnik, D. Oard and P. Bhatia,
    ``Towards Clinical Encounter Summarization: Learning to Compose Discharge Summaries from Prior Notes,''
    \textit{arXiv:2104.13498}.
}
\bibitem{re3writer-model}{
    F. Liu, B. Yang, C. You, X. Wu, S. Ge, Z. Liu, X. Sun, Y. Yang and D. A. Clifton,
    ``Retrieve, Reason, and Refine: Generating Accurate and Faithful Patient Instructions,''
    in \textit{Proc. Neural Inf. Process. Syst.}, 2022.
}
\bibitem{aceso-art-sum}{
    X. Zhang, P. Geng, T. Zhang, Q. Lu, P. Gao and J. Mei,
    ``Aceso: PICO-guided Evidence Summarization on Medical Literature,''
    \textit{IEEE J. Biomed. Health Inform.}, vol. 24, pp. 2663--2670, 2020.
}
\bibitem{sumpubmed}{
    V. Gupta, P. Bharti, P. Nokhiz and H. Karnick,
    ``SumPubMed: Summarization Dataset of PubMed Scientific Articles,''
    in \textit{Proc. Assoc. Comput. Linguistics and Int. Joint Conf. Natural Lang. Process.}, 2021, pp. 292--303.
}
\bibitem{pretrainbiomed-art-sum}{
    Q. Xie, J. A. Bishop, P. Tiwari and S. Ananiadou,
    ``Pre-trained language models with domain knowledge for biomedical extractive summarization,''
    \textit{Knowledge-Based Syst.}, vol. 252. p. 109460, 2022.
}
\bibitem{med-image-cap-baseline}{
    B. Jing, P. Xie and E. Xing,
    ``On the Automatic Generation of Medical Imaging Reports,''
    in \textit{Proc. Assoc. Comput. Linguistics}, 2018, pp. 2577--2586.
}
\bibitem{med-image-cap-KER}{
    C. Y. Li, X. Liang, Z. Hu and E. P. Xing,
    ``Knowledge-driven Encode, Retrieve, Paraphrase for Medical Image Report Generation,''
    in \textit{Proc. AAAI Conf. Artif. Intell.}, 2019, pp. 6666--6673.
}
\bibitem{med-image-cap-topics}{
    H. Nguyen, D. Nie, T. Badamdorj, Y. Liu, Y. Zhu, J. Truong, and L. Cheng,
    ``Automated Generation of Accurate \& Fluent Medical X-ray Reports,''
    in \textit{Proc. Conf. Empirical Methods Natural Lang. Process.}, 2021, pp. 3552--3569.
}
\bibitem{med-image-cap-ppked}{
    F. Liu, X. Wu, S. Ge, W. Fan and Y. Zou,
    ``Exploring and Distilling Posterior and Prior Knowledge for Radiology Report Generation,''
    in \textit{Proc. IEEE Conf. Comput. Vis. Pattern Recognit.}, 2021, pp. 13753--13762.
}
\bibitem{med-image-cap-priorguided}{
    B. Yan, M. Pei, M. Zhao, C. Shan and Z. Tian,
    ``Prior Guided Transformer for Accurate Radiology Reports Generation,''
    \textit{IEEE J. Biomed. Health Inform.}, vol. 26, pp. 5631--5640, 2022.
}
\bibitem{med-image-cap-vlpretrain}{
    J. H. Moon, H. Lee, W. Shin, Y. H. Kim and E. Choi,
    ``Multi-Modal Understanding and Generation for Medical Images and Text via Vision-Language Pre-Training,''
    \textit{IEEE J. Biomed. Health Inform.}, vol. 26, pp. 6070--6080, 2022.
}
\bibitem{ai-review-sum-survey}{
    M. Gambhir and V. Gupta,
    ``Recent automatic text summarization techniques: a survey,''
    \textit{Artif. Intell. Review}, vol. 47, pp. 1--66, 2017.
}
\bibitem{aaai-sum-survey}{
    H. Lin and V. Ng,
    ``Abstractive Summarization: A Survey of the State of the Art,''
    in \textit{Proc. AAAI Conf. Artif. Intell.}, 2019, pp. 9815--9822.
}
\bibitem{eswa-sum-survey}{
    W. S. El-Kassas, C. R. Salama, A. A. Rafea and H. K. Mohamed,
    ``Automatic text summarization: A comprehensive survey,''
    \textit{Expert Systems with Applications}, vol. 165, p. 113679, 2021.
}
\bibitem{transformer-model}{
    A. Vaswani, N. Shazzer, N. Parmar, J. Uszkoreit, L. Jones, A. N. Gomez, Ł. Kaiser and I. Polosukhin, 
    ``Attention is all you need,'' 
    in \textit{Proc. Neural Inf. Process. Syst.}, 2017, pp. 5998--6008.
}
\bibitem{bert-model}{
    J. Devlin, M. Chang, K. Lee, and K. Toutanova, 
    ``BERT: Pre-training of deep bidirectional Transformers for language understanding,'' 
    in \textit{Proc. North. Amer. Chapter Assoc. Comput. Linguistics}, 2019, pp. 4171--4186.
}
\bibitem{bart-model}{
    M. Lewis, Y. Liu, N. Goyal, M. Ghazvininejad, A. Mohamed, O. Levy, V. Stoyanov, and L. Zettlemoyer,
    ``BART: Denoising Sequence-to-Sequence Pre-training for Natural Language Generation, Translation, and Comprehension,'' 
    in \textit{Proc. Assoc. Comput. Linguistics}, 2020, pp. 7871--7880.
}
\bibitem{t5-model}{
    C. Raffel, N. Shazeer, A. Roberts, K. Lee, S. Narang, M. Matena, Y. Zhou, W. Li and P. J. Liu, 
    ``Exploring the Limits of Transfer Learning with a Unified Text-to-Text Transformer,'' 
    \textit{arXiv:1910.10683}.
}
\bibitem{unilm-model}{
    L. Dong, N. Yang, W. Wang, F. Wei, X. Liu, Y. Wang, J. Gao, M. Zhou and H. W. Hon,
    ``Unified Language Model Pre-training for Natural Language Understanding and Generation,''
    in \textit{Proc. Neural Inf. Process. Syst.}, 2019, pp. 13042--13054.
}
\bibitem{bertabsext}{
    Y. Liu and M. Lapata,
    ``Text Summarization with Pretrained Encoders,''
    in \textit{Proc. Conf. Empirical Methods Natural Lang. Process. and Int. Joint Conf. Natural Lang. Process.}, 2019, pp. 3730-3740.
}
\bibitem{pegasus-model}{
    J. Zhang, Y. Zhao, M. Saleh and P. Liu,
    ``PEGASUS: pre-training with extracted gap-sentences for abstractive summarization,''
    in \textit{Proc. Int. Conf. Mach. Learn.}, 2020, pp. 11328--11339.
}
\bibitem{prophetnet-model}{
    W. Qi, Y. Yan, Y. Gong, D. Liu, N. Duan, J. Chen, R. Zhang and M. Zhou,
    ``ProphetNet: Predicting Future N-gram for Sequence-to-SequencePre-training,''
    in \textit{Proc. Findings Assoc. Comput. Linguistics: Empirical Methods Natural Lang. Process.} 2020, pp. 2401--2410.
}
\bibitem{post-edit-correct-fact}{
    M. Cao, Y. Dong, J. Wu, and J. C. K. Cheung,
    ``Factual Error Correction for Abstractive Summarization Models,''
    in \textit{Proc. Conf. Empirical Methods Natural Lang. Process.}, 2020. pp. 6251--6258.
}
\bibitem{fes-model}{
    X. Chen, M. Li, X. Gao and X. Zhang,
    ``Towards Improving Faithfulness in Abstractive Summarization,''
    in \textit{Proc. Neural Inf. Process. Syst.}, 2022.
}
\bibitem{factpegasus-model}{
    D. Wan and M. Bansal,
    ``FactPEGASUS: Factuality-Aware Pre-training and Fine-tuning for Abstractive Summarization,''
    in \textit{Proc. North. Amer. Chapter Assoc. Comput. Linguistics: Human language Technologies}, 2022, pp. 1010--1028.
}
\bibitem{gsum-model}{
    Z.-Y. Dou, P. Liu, H. Hayashi, Z. Jiang, and G. Neubig, 
    ``GSum: A General Framework for Guided Neural Abstractive Summarization,'' 
    in \textit{Proc. North. Amer. Chapter Assoc. Comput. Linguistics: Human language Technologies}, 2021, pp. 4830--4842.
}
\bibitem{salience-ext-alloc-model}{
    F. Wang, K. Song, H. Zhang, L. Jin, S. Cho, W. Yao, X. Wang, M. Chen, and D. Yu,
    ``Salience Allocation as Guidance for Abstractive Summarization,''
    in \textit{Proc. Conf. Empirical Methods Natural Lang. Process.}, 2022, pp. 6094--6106.
}
\bibitem{rouge-metric}{
    C. Y. Lin,
    ``ROUGE: A Package for Automatic Evaluation of Summaries,''
    in \textit{Proc. Assoc. Comput. Linguistics}, 2004, pp. 74--81.
}
\bibitem{bertscore-metric}{
    T. Zhang and V. Kishore, F. Wu, K. Q. Weinberger and Y. Artzi,
    ``BERTScore: Evaluating Text Generation with BERT,''
    in \textit{Proc. Int. Conf. Learn. Representations}, 2020.
}
\bibitem{bartscore-metric}{
    W. Yuan, G. Neubig and P. Liu,
    ``BARTScore: Evaluating Generated Text as Text Generation,''
    in \textit{Proc. Neural Inf. Process. Syst.}, 2021, pp. 27263--27277.
}
\bibitem{summac-metric}{
    P. Laban, T. Schnabel, P. N. Bennett and M. A. Hearst,
    ``SummaC: Re-Visiting NLI-based Models for Inconsistency Detection in Summarization,'' 
    \textit{Trans. Assoc. Comput. Linguistics}, vol. 10, pp. 163--177, 2022.
}
\bibitem{questeval-metric}{
    T. Scialom, P. A. Dray, S. Lamprier, B. Piwowarski, J. Staiano, A. Wang, and P. Gallinari,
    ``QuestEval: Summarization Asks for Fact-based Evaluation,''
    in \textit{Proc. Conf. Empirical Methods Natural Lang. Process.}, 2021, pp. 6594--6604.
}
\bibitem{factcc-metric}{
    W. Kryscinski, B. McCann, C. Xiong, and R. Socher,
    ``Evaluating the Factual Consistency of Abstractive Text Summarization,''
    in \textit{Proc. Conf. Empirical Methods Natural Lang. Process.}, 2022, pp. 9332--9346.
}
\bibitem{clinical-sum-sys}{
    H. Moen, L. M. Peltonen, J. Heimonen, A. Airola, Y. Pahikkala, T. Salakoski and S. Salanterä,
    ``Comparison of automatic summarisation methods for clinical free text notes,''
    \textit{Artif. Intell. in Medicine}, vol. 67, pp. 25-37, 2016.
}
\bibitem{ehr-sum-sys}{
    S. H. Lee,
    ``Natural language generation for electronic health records,''
    \textit{npj digital medicine}, vol. 1, 2018.
}
\bibitem{standford-sum-opt-fact}{
    Y. Zhang, D. Merck, E. Tsai, C. D. Manning and C. Langlotz,
    ``Optimizing the Factual Correctness of a Summary: A Study of Summarizing Radiology Reports,''
    in \textit{Proc. Assoc. Comput. Linguistics}, 2020, pp. 5108--5120.
}
\bibitem{mimic-iii}{
    A. E. W. Johnson, T. J. Pollard, and R. G. Mark, 
    ``MIMIC-III clinical database (version 1.4),'' 
    \textit{PhysioNet}, 2020. DOI: \url{https://doi.org/10.13026/C2XW26}.
}
\bibitem{mimic-iii-paper}{
    A. E. W. Johnson, T. J. Pollard, L. Shen, L. H. Lehman, M. Feng, M. Ghassemi, B. Moddy, P. Szolovits, L. A. Ceil and R. G. Mark, 
    ``MIMIC-III, a freely accessible critical care database,'' 
    \textit{Sci. Data}, vol. 3, no. 1, p. 160035, 2016.
}
\bibitem{gap-statistic}{
    A. Cuevas, M. Febrero. and R. Fraiman, 
    ``Estimating the Number of Clusters,'' 
    \textit{The Cana. J. of Stat. / La Revue Canadienne de Statistique}, 28(2), pp. 367--382, 2000.
}
\bibitem{k-means}{
    D. Arthur and S. Vassilvitskii, 
    ``K-means++: the advantages of careful seeding,'' 
    in \textit{Proc. ACM-SIAM symp. on Disc. algo. (SODA)}, pp. 1027--1035, 2007.
}
\bibitem{tsne-visualize}{
    L.V.D. Maaten and G. Hinton, 
    ``Visualizing data using t-sne,'' 
    \textit{J. Mach. Learn. Res.}, vol. 9, no. Nov, pp. 2579--2605, 2008.
}
\bibitem{silhouette-score}{
    P. J. Rousseeuw, 
    ``Silhouettes: a Graphical Aid to the Interpretation and Validation of Cluster Analysis,'' 
    \textit{Comput. and Applied Math.}, vol. 20, pp. 53--65, 1987.
}
\bibitem{adamW}{
    I. Loshchilov and F. Hutter, 
    ``Decoupled Weight Decay Regularization,'' 
    in \textit{Proc. Int. Conf. Learn. Representations}, 2019.
}

\end{thebibliography}
\end{document}